\documentclass[11pt]{article}

\usepackage[final]{acl}

\usepackage{times}
\usepackage{latexsym}

\usepackage[T1]{fontenc}

\usepackage[utf8]{inputenc}

\usepackage{microtype}

\usepackage{inconsolata}

\usepackage{graphicx}

\usepackage{algorithm}
\usepackage{algorithmic}
\usepackage{booktabs}  
\usepackage{multirow}  
\usepackage{wrapfig}
\usepackage{subcaption}     
\usepackage{amssymb} 
\usepackage{amsmath}

\usepackage{xcolor}
\definecolor{commentcolor}{HTML}{067f7f}

\title{Structuring Reasoning for Complex Rules Beyond Flat Representations}

\author{
  \textbf{Zhihao Yang\textsuperscript{1,2}\thanks{Equal contribution}},
  \textbf{Ancheng Xu\textsuperscript{1,2}\footnotemark[1]},
  \textbf{Jingpeng Li\textsuperscript{5}},
  \textbf{Liang Yan\textsuperscript{5}},
  \textbf{Jiehui Zhou\textsuperscript{5}}, \\
  \textbf{Zhen Qin\textsuperscript{5}},
  \textbf{Hengyu Chang\textsuperscript{5}},
  \textbf{Yukun Chen\textsuperscript{1,2}},
  \textbf{Longze Chen\textsuperscript{1,2}},
  \textbf{Ahmadreza Argha\textsuperscript{3}},\\
  \textbf{Hamid Alinejad-Rokny\textsuperscript{3}}, 
  \textbf{Minghuan Tan\textsuperscript{1}}, 
  \textbf{Yujun Cai\textsuperscript{4}\thanks{Corresponding authors}},
  \textbf{Min Yang\textsuperscript{1}\footnotemark[2]}, \\
  \textsuperscript{1}Shenzhen Key Laboratory for High Performance Data Mining, \\
  Shenzhen Institutes of Advanced Technology, Chinese Academy of Sciences \\
  \textsuperscript{2}University of Chinese Academy of Sciences,
  \textsuperscript{3}School of Biomedical Engineering, UNSW Sydney \\
  \textsuperscript{4}University of Queensland,
  \textsuperscript{5}Alibaba Group \\
  \texttt{zh.yang30@outlook.com, min.yang@siat.ac.cn}
}

\begin{document}
\maketitle
\begin{abstract}
Large language models (LLMs) face significant challenges when processing complex rule systems, as they typically treat interdependent rules as unstructured textual data rather than as logically organized frameworks.
This limitation results in reasoning divergence, where models often overlook critical rule dependencies essential for accurate interpretation. Although existing approaches such as Chain-of-Thought (CoT) reasoning have shown promise, they lack systematic methodologies for structured rule processing and are particularly susceptible to error propagation through sequential reasoning chains.
To address these limitations, we propose the Dynamic Adjudication Template (DAT), a novel framework inspired by expert human reasoning processes. DAT structures the inference mechanism into three methodical stages: \emph{qualitative analysis, evidence gathering, and adjudication}. During the \emph{qualitative analysis} phase, the model comprehensively evaluates the contextual landscape. The subsequent \emph{evidence gathering} phase involves the targeted extraction of pertinent information based on predefined template elements ([placeholder]), followed by systematic verification against applicable rules. Finally, in the \emph{adjudication} phase, the model synthesizes these validated components to formulate a comprehensive judgment.
Empirical results demonstrate that DAT consistently outperforms conventional CoT approaches in complex rule-based tasks. Notably, DAT enables smaller language models to match, and in some cases exceed, the performance of significantly larger LLMs, highlighting its efficiency and effectiveness in managing intricate rule systems.
\end{abstract}

\begin{figure}[t]
    \centering
    \includegraphics[width=1\columnwidth]{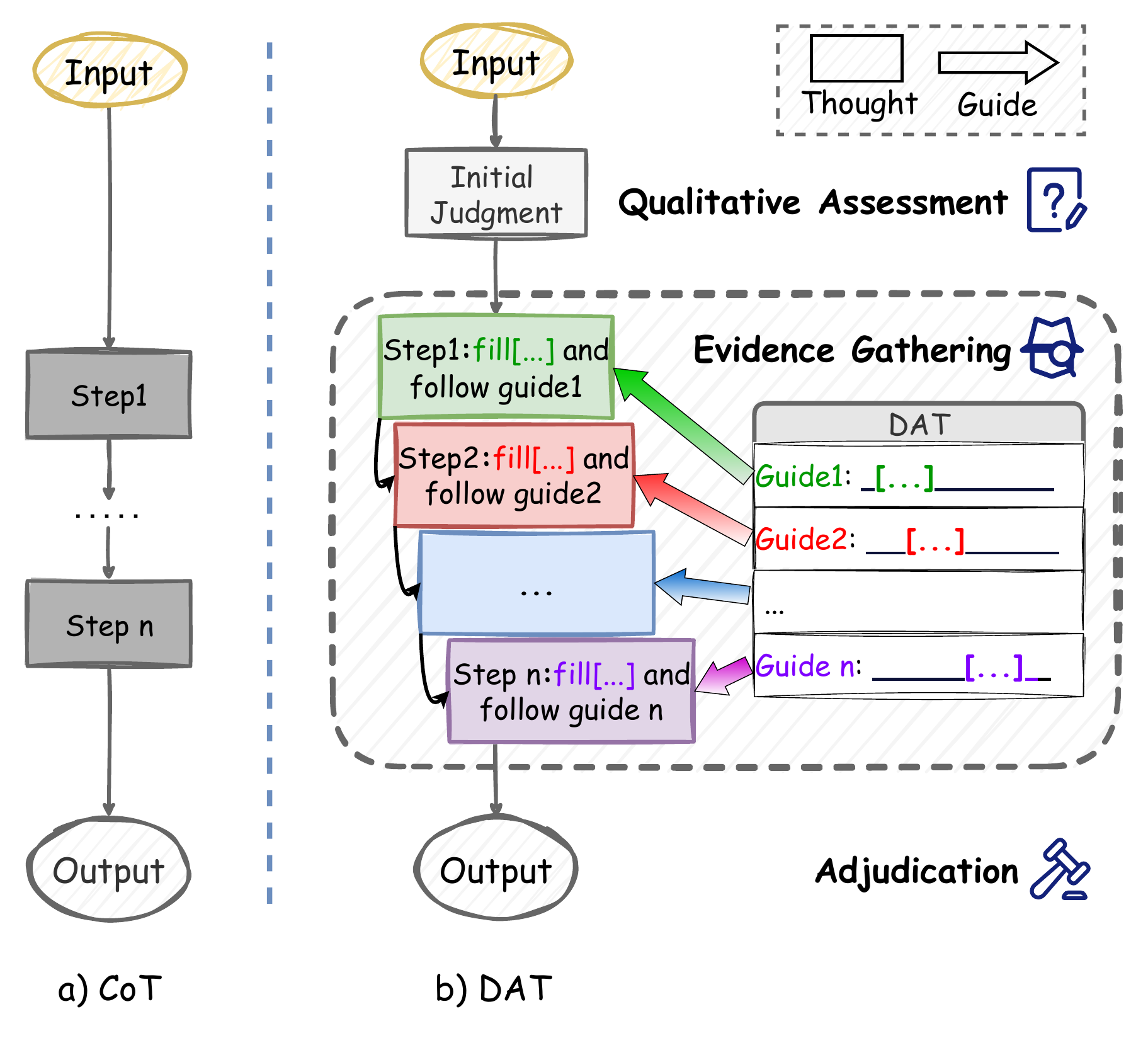}
    \caption{Illustration of different reasoning process. Dynamic Adjudication Template enables large language models to organizes the inference mechanism into a structured three-stage process.}
    \label{fig:dat}
\end{figure}

\section{Introduction}
Large language models (LLMs)~\cite{zhao2023survey,Xiao_2025} have exhibited remarkable capabilities across a wide range of natural language processing tasks~\cite{vaswani2017attention,brown2020language}, demonstrating significant potential in rule-intensive domains such as e-commerce content moderation~\cite{li2024ad}, legal advisory services~\cite{yang2024large}, and financial risk analysis~\cite{wu2023bloomberggpt}. These domains demand precise rule interpretation and rigorous implementation, as even minor errors in judgment can result in legal risks, financial losses, or serious damage to credibility.
However, when processing complex instruction sets characterized by densely interdependent rules and nuanced semantic relationships~\cite{ouyang2022training}, the performance of LLMs deteriorates significantly~\cite{srivastava2023beyond}. 
For instance, when an advertisement must simultaneously convey the seller's promotional message and strictly comply with advertising laws, LLMs often struggle to navigate the subtle interactions between overlapping rules, leading to inconsistent judgments~\cite{xu2025evade}. 
This limitation stems from two fundamental problems: (1) \textbf{Flat rule processing}, where models treat rule systems as unstructured text collections rather than hierarchically organized frameworks with explicit logical relationships; and (2) \textbf{Reasoning divergence}, where models are distracted by superficial features or misled by ambiguous cues~\cite{ji2023survey}, failing to identify the most relevant rules.

Existing reasoning approaches have attempted to address complex reasoning challenges
but fall short in rule-intensive scenarios. Chain-of-Thought (CoT)~\cite{wei2022chain} and its variants fail to ensure that each step strictly adheres to logic and facts. Errors in any intermediate step can propagate and collapse the coherence of the entire reasoning chain~\cite{lanham2023measuring}. Tree-based methods like Tree-of-Thought (ToT)~\cite{yao2023tree} explore multiple paths but lack systematic rule verification. These
approaches share a common limitation: they lack mechanisms for hierarchical rule organization, structured evidence verification, and systematic conflict resolution when multiple rules apply simultaneously.

This gap reveals a fundamental mismatch between current LLM reasoning approaches and the requirements of rule-intensive domains. Effective rule-based reasoning calls for a new framework that enables structured verification processes rather than relying on free-form exploration. To this end, we propose a novel method, the \textbf{Dynamic Adjudication Template (DAT)}, illustrated in \textbf{Figure~\ref{fig:dat}.} This method emulates the cognitive approach of human experts~\cite{liao2021human}. Rather than beginning with detailed computation, experts first construct a high-level problem-solving framework and then identify critical points for focused analysis.
Unlike static prompting techniques that rely on fixed templates, DAT introduces a dynamic, structured reasoning paradigm. It actively guides the model through a structured three-step process of \emph{Qualitative Assessment, Evidence Gathering, and Adjudication.} 
The process begins with a high-level assessment of the problem. The model then focuses on key, complex, and error-prone decision points most relevant to the question, enabling targeted analysis. These targeted insights are finally synthesized into a logically sound and comprehensive adjudication. 
In our experiments on a rule-intensive e-commerce dataset, DAT improved the overall accuracy of Qwen-2.5-7B~\cite{qwen2025qwen25technicalreport} from 34.11\% to 62.49\%. It also outperformed larger CoT-equipped models such as Qwen-Max and Deepseek-R1~\cite{guo2025deepseek} on several complex rule-based subtasks within the dataset.

In summary, our contributions are threefold.
\begin{enumerate}
    \item We propose the DAT approach, a human-inspired strategy that guides models through a structured three-step process: \emph{Qualitative Assessment, Evidence Gathering, and Adjudication}. DAT transforms the reasoning paradigm from flat rule processing to hierarchical reasoning, and from uncontrolled error propagation to structured verification—enabling more coherent and rule-consistent judgments in complex tasks.

    \item We introduce an automated pipeline for template generation, filtering, and selection. By replacing static prompts with adaptive templates, we enable flexible and context-aware reasoning across diverse rule scenarios. This design ensures both interpretability and adaptability, supporting controlled and effective reasoning in a wide range of rule-based tasks.

    \item We report substantial gains on rule-intensive benchmarks. Our method enables small, efficient models to outperform much larger LLMs using Chain-of-Thought reasoning on complex rule-based tasks. This paves the way for high-performance applications in low-resource settings. Preliminary results also suggest promising generalization to vision-language models (VLMs)~\cite{liang2024comprehensive,li2024multimodal}, indicating a fruitful direction for future research.

\end{enumerate}

\begin{algorithm*}[t]
   \caption{Three-Stage Structured Reasoning Pipeline}
   \label{alg:algorithm1}
 \textbf{Input}: Question $q$, templates $T_K$, rules $\mathcal{R}$\\
 \textbf{Parameter}: Template manager $\mathcal{M}$ \\
 \textbf{Output}: Judgment $\hat{J}$
   \begin{algorithmic}[1]
   \STATE \textbf{Step 1: Template Selection} 
   \STATE $t^* \gets \mathcal{M}(q, T_K)$ \textcolor{commentcolor}{// Select optimal template for the query.}
   
   \STATE \textbf{Step 2: Initial Judgment \& Evidence Extraction}
   \STATE $J_\text{initial} \leftarrow \text{QualitativeAnalysis}(q, t^*)$ \textcolor{commentcolor}{// Form an initial, holistic judgment.}
   \STATE $\mathcal{P} \leftarrow \text{GetPlaceholders}(t^*)$ \textcolor{commentcolor}{// Extract key placeholders from the template.}
   \FOR{each $p \in \mathcal{P}$}
       \STATE $\mathcal{E}_p \gets f_\text{extract}(p, q, \mathcal{R})$ \textcolor{commentcolor}{// Extract evidence for each placeholder.}
       \STATE $\mathcal{V}_p \gets f_\text{match}(\mathcal{E}_p, \mathcal{R})$ \textcolor{commentcolor}{// Match the evidence against rules.}
   \ENDFOR
   
   \STATE \textbf{Step 3: Evidence Integration}
   \STATE Assemble evidence chain $\mathcal{C} \gets \{\mathcal{V}_{p}\}_{p\in\mathcal{P}}$ \textcolor{commentcolor}{// Assemble validated evidence.}
   \STATE $\hat{J} \gets f_\text{adjudicate}(J_\text{initial}, \mathcal{C})$ \textcolor{commentcolor}{// Adjudicate final judgment with evidence.}
   
   \STATE \textbf{Return} $\hat{J}$
   \end{algorithmic}
\end{algorithm*}

\section{Related Work}
\subsection{Complex Rule-based Reasoning Benchmark} 
LLMs and VLMs have been widely adopted in e-commerce applications such as content moderation, product recommendation, and search~\cite{jiang2025beyond,palen2024investigating}. Nevertheless, enabling these models to make accurate judgments under complex and dynamic platform rules remains a significant challenge. To evaluate model capabilities for such tasks, both academia and industry have developed specialized benchmarks. Among these, EVADE stands out as the first multimodal benchmark for evasive content detection in Chinese e-commerce. Derived from authentic advertising regulations and annotated by domain experts, this dataset comprises 2,833 textual samples and 13,961 images across six product categories. EVADE evaluates model performance via two distinct tasks: Single-Violation Judgment and All-in-One Judgment. By embedding semantically overlapping rules within instructions, it requires reasoning over complex, long-context, rule-dense scenarios, closely aligning with our focus on complex rule interpretation. Research using EVADE has revealed limitations in current models' capacity to comprehend and apply multi-layered rules, highlighting the necessity for our proposed approach.

\begin{figure*}[t] 
    \centering
    \includegraphics[width=\textwidth]{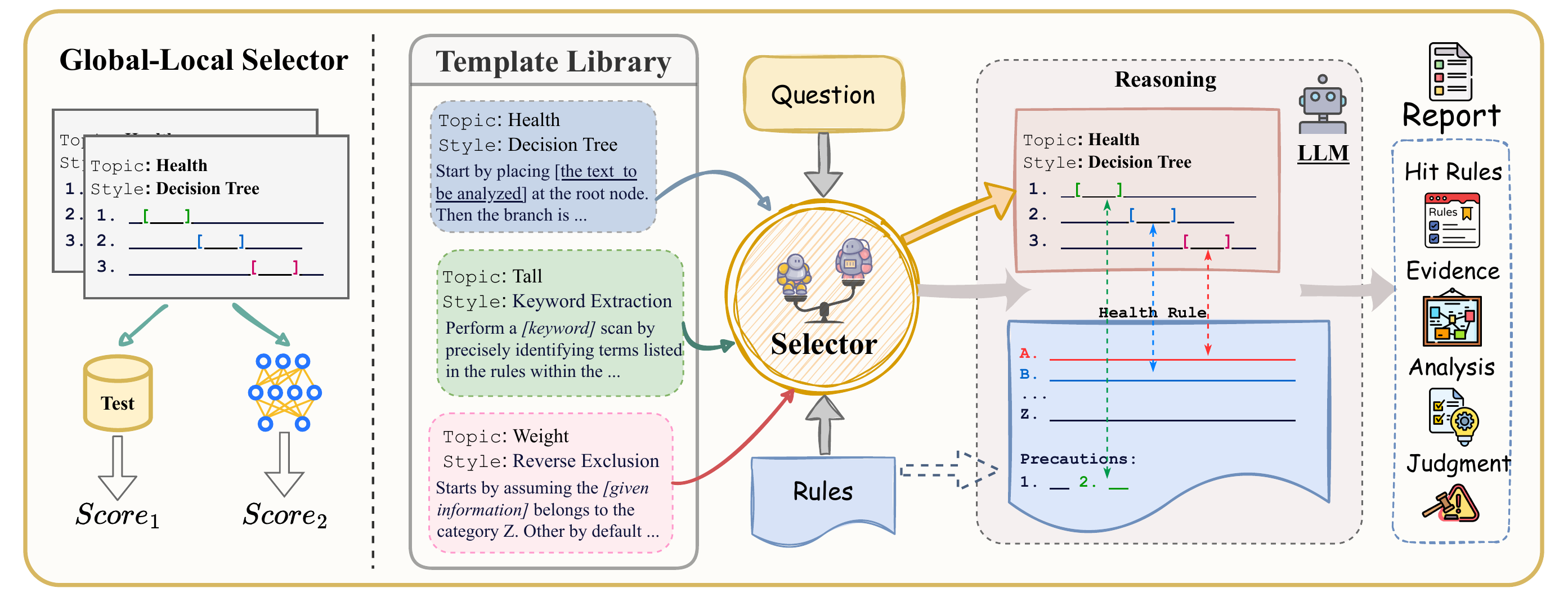} 
    \caption{(Left) The architecture of the Global-Local Selector. (Right) When processing complex rule systems, LLM utilizes the selected DAT through Qualitative Analysis, Evidence Gathering, and Adjudication to generate a comprehensive judgment.}
    \label{fig:dat_framework}
\end{figure*}

\subsection{Prompt-based Reasoning Methods with Large Language Models}
To advance the reasoning capabilities of large language models, researchers have developed numerous prompt-based methodologies. Prominent among these are CoT prompting~\cite{wei2022chain} and its derivatives, including Least-to-Most~\cite{zhou2022least} and Decomposed Prompting~\cite{khot2022decomposed}, which decompose complex problems into sequential subtasks and have proven effective across diverse reasoning domains. However, the inherent structural fragility of linear reasoning approaches manifests as cascading error propagation during sequential processing stages, as demonstrated in~\citet{yang2024buffer}. Emerging paradigms such as Tree-of-Thought and Graph-of-Thought~\cite{besta2024graph} mitigate this issue through adaptive, non-linear heuristic search mechanisms. While these frameworks offer potential for improved reasoning fidelity, they introduce substantial computational costs and rely on handcrafted, domain-specific prompting strategies~\cite{yang2024buffer}.
This rigid templating paradigm is particularly deficient for dynamic, evolving real-world environments. Within our problem domain, the governing rules exhibit both combinatorial complexity and temporal adaptation, rendering rule-category-specific manual prompt engineering impractical and misaligned with the dynamic nature of the challenge.

\section{Methodology}
In this section, we introduce our novel framework for rule-intensive applications. 
For a given query \(Q\) and rules \(\mathcal{R}\), the framework selects the optimal reasoning template \(t^*\) from a predefined library \(T_K\), which then guides the model in executing a structured reasoning process, as shown in \textbf{Figure~\ref{fig:dat_framework}.} To address the challenges of template generation, selection, and application, we propose a pipeline framework that consists of three primary components:
1) \textbf{Three-Stage Structured Reasoning.} This component serves as the core of our framework. It utilizes the selected template to execute a structured three-step reasoning process—\emph{Qualitative Analysis, Evidence Gathering, and Adjudication}—to derive a logically grounded and rule-consistent result. This structured inference relies on high-quality, task-aligned templates, which are sourced and selected through the following two components.
2) \textbf{Dynamic Template Library Construction.} To support the reasoning process, this component systematically generates and validates a diverse repository of high-quality reasoning templates, which serve as the foundational knowledge base for the framework.
3) \textbf{Adaptive Template Selection.} Given a query, this component dynamically selects the optimal template from the library by evaluating both general performance and task-specific fitness. Detailed prompts for each component can be found in the \textbf{appendix~\ref{template_prommpt}.}

\subsection{Three-Stage Structured Reasoning}
To avoid reasoning divergence and error propagation in multi-step inference, we employ a structured, template-guided approach. As indicated in \textbf{Algorithm~\ref{alg:algorithm1}} and the right side of \textbf{Figure~\ref{fig:dat_framework}}, when encountering complex rules during reasoning, the model employs a dynamic template manager, which selects the appropriate template for the given task and question from a curated template knowledge base. The model then executes a structured three-step reasoning process: \emph{Qualitative Analysis, Evidence Gathering, and Adjudication}.

First, \emph{Qualitative Analysis} guides the model to review all relevant information and form an initial, holistic judgment. This step prioritizes broad contextual understanding over premature attention to isolated details. Next, \emph{Evidence Gathering} utilizes predefined [placeholders] within the selected template, representing \textbf{key reasoning checkpoints, complex decision nodes, and error-prone components,} to extract task-specific information from the question. The retrieved information is independently matched against relevant rules to verify consistency and correctness. Finally, \emph{Adjudication} instructs the model to re-evaluate its initial judgment using the validated \textbf{evidence-rule chain}, producing a final decision through structured and rigorous logical reasoning.
To support this reasoning process, we next describe how to construct a high-quality, dynamic template library.

\subsection{Template Library Construction}
Our goal is to develop an efficient pipeline for generating data. This pipeline will build a diverse, high-quality, dynamically extensible library of reasoning templates. This repository will enable LLMs to adaptively select appropriate templates during their reasoning processes. The complete procedure is formalized in \textbf{Algorithm~\ref{alg:algorithm_dat_library}.}

\begin{algorithm*}[tb]
    \caption{DAT Library Construction Pipeline}
    \label{alg:algorithm_dat_library}
    \textbf{Input}: Task context, full dataset $D$ \\
    \textbf{Parameters}: \\
    \quad Seed template count $m$, Prefix length $k$, Style Count $v$, \\
    \quad Sample ratio $r$ (default: 0.2), Performance threshold $\theta$ \\
    \textbf{Output}: Refined template set $T_K$, selector training data $S$
    
    \begin{algorithmic}[1]
    
    \STATE \textbf{Step 1: Seed Template Generation}
    \STATE $T_0 \leftarrow \text{GenerateSeeds}(\text{context}, m)$ \textcolor{commentcolor}{// Generate m initial seed templates.}
    
    \STATE \textbf{Step 2: Template Expansion \& Augmentation}
    \STATE $T_1 \leftarrow \text{ExpandWithPrefix}(T_0, k, \text{LLM})$ \textcolor{commentcolor}{// Expand seeds via structured continuation.}
    \STATE $T_2 \leftarrow \text{StyleTransfer}(T_1, v)$ \textcolor{commentcolor}{// Apply v style variants to each template.}
    
    \STATE \textbf{Step 3: Template Evaluation}
    \STATE $D_1 \leftarrow \text{Sample}(D, r)$ \textcolor{commentcolor}{// Randomly sample a subset $D_1$ for evaluation.}
    \STATE $T_K \leftarrow \{ t \in T_2 \mid \text{score}(t, D_1) \geq \theta \}$ \textcolor{commentcolor}{// Filter templates based on performance score.}

    \STATE \textbf{Step 4: Training Data Generation}
    \STATE $S \leftarrow \{ (\text{score}(t, d), t, d) \mid t \in T_K, d \in D_1 \}$ \textcolor{commentcolor}{// Construct training set S with scores for selector.}
    
    \STATE \textbf{Return} $T_K$, $S$ \textcolor{commentcolor}{// Output the refined library and training data.}

    \end{algorithmic}
\end{algorithm*}

\subsubsection{Template Generation and Expansion}
We introduce a systematic template generation pipeline using Gemini 2.5-Pro~\cite{comanici2025gemini}. The process begins by creating 10 distinct seed templates ($T_0$) using only task context, without problem-specific details. These templates capture diverse analytical perspectives.
To enhance coverage and robustness, we perform multi-stage augmentation through structured continuation~\cite{zhou2022large,madaan2023self} and style transfer~\cite{jin2022deep}. 
Structured continuation begins by providing the model with the first several steps of each seed template—specifically, from step 1 to step $n{-}1$. The model is then instructed to complete the remaining steps in a consistent and logical manner, expanding $T_0$ into a structurally diverse set $T_1$.
We then apply style transfer to $T_1$ to introduce linguistic and syntactic diversity.
By not specifying the exact styles, we harness the model's inherent understanding of language to produce a diverse and less biased set of augmented data.
Specifically, the model is instructed to modify existing templates in $v$ distinct styles. 
This step modifies expressions while preserving the task relevance, yielding the final augmented set $T_2$ with improved stylistic diversity and task adaptability.

Template effectiveness is dynamically evaluated on a randomly sampled 20\% subset of the data ($D_1 \subset D$), which serves as an approximate evaluation set for estimating generalizability. We assess each template in $T_2$ based on its ability to generate accurate and complete outputs for examples in $D_1$, and filter those that show reliable performance across varied examples in $D_1$, forming the refined library $T_K$. Performance records from applying $T_K$ to $D_1$ are then used to construct a training dataset $S$ for the template selector.

\subsubsection{Partitioning an Independent Dataset}
We construct an independent evaluation dataset $D_1$ by randomly sampling 20\% of the complete dataset. This strategically sampled subset serves two key purposes in our pipeline.
First, $D_1$ enables performance evaluation and filtering of the template library $T_2$. We measure critical metrics including error rates and task-specific accuracy across all templates in $T_2$. This assessment identifies high-performing templates that are optimally aligned with the task context, yielding a refined subset $T_K$.
Second, $D_1$ generates training data for the template selector. We record the performance metrics of $T_K$'s predictions on $D_1$ to construct a supervised training dataset, which supports training the template selector discussed in Section Local Selector $S_2$. This enables context-aware, dynamic template selection across diverse task scenarios.
With the template library constructed, we next introduce an adaptive selection mechanism to choose the most suitable template for each query.

\begin{table*}[htbp]
    \small
    \renewcommand{\arraystretch}{1.3} 
    \setlength{\tabcolsep}{10pt} 
    \centering
    \begin{tabular}{l l|c c c c c c c}
        \toprule[1.5pt] 
        \textbf{Model} & \textbf{Method} & \textbf{Overall} & \textbf{Body} & \textbf{Women} & \textbf{Height} & \textbf{Men} & \textbf{Weight} & \textbf{Health} \\
        \midrule
        \multicolumn{9}{c}{\textbf{Small Models (Baseline \& Ours)}} \\
        \hline
        \multirow{2}{*}{ChatGLM3-6B} 
        & CoT & 25.39 & 43.83 & 42.60 & 24.60 & 30.84 & 21.19 & 4.68 \\
        & Ours & 43.82 {\scriptsize ($\uparrow$18.43)} & 67.28 & 62.72 & 39.73 & 48.28 & 29.94 & 23.26 \\
        \hline
        \multirow{2}{*}{ChatGLM4-9B} 
        & CoT & 37.54 & 30.25 & 42.60 & 29.80 & 53.26 & 57.06 & 28.27 \\
        & Ours & 59.57 {\scriptsize ($\uparrow$22.03)} & 63.58 & \textbf{79.88} & 54.40 & 63.41 & 73.45 & 38.93 \\
        \hline
        \multirow{2}{*}{InternLM3-8B} 
        & CoT & 45.54 & 39.51 & 37.28 & 52.82 & 61.88 & 51.69 & 39.58 \\
        & Ours & 56.71 {\scriptsize ($\uparrow$11.17)} & 58.64 & 64.50 & 59.82 & 63.79 & 65.25 & 40.23 \\
        \hline
        \multirow{2}{*}{Qwen-2.5-7B} 
        & CoT & 34.11 & 24.69 & 37.87 & 41.08 & 53.83 & 38.14 & 29.56 \\
        & Ours & \textbf{62.49 {\scriptsize ($\uparrow$28.38)}} & \textbf{77.78} & 78.11 & \textbf{61.40}& 62.84 & 61.30 & \textbf{41.36} \\
        \hline
        \multirow{2}{*}{Qwen-2.5-14B} 
        & CoT & 45.49 & 41.36 & 56.80 & 48.98 & 50.00 & 57.06 & 33.12 \\
        & Ours & 58.16 {\scriptsize ($\uparrow$12.67)} & 58.64 & 76.92 & 53.27 & \textbf{65.71} & \textbf{73.45} & 39.58 \\
        \hline
        \multicolumn{9}{c}{\textbf{Large Models (Baseline Only)}} \\
        \hline
        Qwen-2.5-72B & CoT & 51.16 & 62.35 & 61.54 & 40.86 & 57.09 & 58.76 & 33.28 \\
        Qwen-max & CoT & 51.73 & 64.20 & 67.46 & \underline{45.60} & \underline{58.05} & 55.65 & 30.86 \\
        GPT-4.1-0414 & CoT & 53.18 & 53.70 & 71.01 & 44.70 & 55.94 & 73.73 & \underline{34.41} \\
        Deepseek-R1 & CoT & \underline{60.87} & \underline{74.69} & \underline{77.51} & 43.57 & 55.56 & \underline{74.58} & 41.03 \\
        \bottomrule[1.5pt] 
    \end{tabular}
    \caption{Performance Comparison of LLM Models on Partial Accuracy. Bold values represent the best performance across small parameter models, while underlined values indicate the best performance across large parameter models.}
    \label{table:model_performance_main_result}
\end{table*}

\subsection{Adaptive Template Selection}
In rule-intensive scenarios, ensuring absolute controllability and determinism is critical. Directly using large-model-generated reasoning templates risks violating domain-specific constraints~\cite{huang2025survey}. To address this, we propose a Global-Local Template Selector that draws from a curated repository of pre-validated templates~\cite{lewis2020retrieval}, ensuring constraint compliance and process reliability. As shown on the left of \textbf{Figure~\ref{fig:dat_framework},} the selector retrieves and validates templates to align with rule-intensive constraints. Our dual-evaluation framework assesses both the \textbf{macro-level stability across tasks} and \textbf{adaptability to individual instances}. We identify optimal templates through a weighted integration of global and local performance metrics~\cite{john2010elements}, with min-max normalization applied to both scores:

{\small
\begin{equation}
\begin{aligned}
s_{\text{final}}(T_i) = \lambda \cdot \frac{s_1(T_i) - \min_j s_1(T_j)}{\max_j s_1(T_j) - \min_j s_1(T_j)} \\
+ (1 - \lambda) \cdot \frac{s_2(T_i) - \min_j s_2(T_j)}{\max_j s_2(T_j) - \min_j s_2(T_j)}.
\end{aligned}
\end{equation}
}
where $s_1$ denotes the global performance across tasks, $s_2$ the local fitness to the current query, and $\lambda \in [0,1]$ a tunable hyperparameter balancing global stability and local fit.

\subsubsection{Global Selector $S_1$}  
It evaluates templates based on their accuracy on $D_1$, reflecting general reliability across tasks. The global score $s_1$ is defined as:  
\begin{equation}
\ s_1(T_i) = \operatorname{Acc}(T_i, D_1).
\end{equation}
Here, $T_i$ represents each template in the library, and $Acc(T_i, D_1)$ denotes the prediction accuracy of the template $T_i$ on the dataset $D_1$. 

\subsubsection{Local Selector $S_2$} 
To construct the training dataset, we pair reasoning templates that produce correct solutions with those that yield incorrect solutions for the same queries. Specifically, when template $T_A$ generates a correct result for a query and template $T_B$ generates an incorrect result, $T_A$ is considered superior to $T_B$ for that query. Using this method, we randomly sample 12,000 correct-incorrect template pairs from each of the six risk categories for training~\cite{rafailov2023direct}.  
\begin{multline}
D_{\text{train}} = \{ (T_A, T_B, Q) \mid Q \in D_1,\;
T_A, T_B \in T_K, \\
\mathbb{I}(R(T_A, Q)) = 1,\;
\mathbb{I}(R(T_B, Q)) = 0 \}
\end{multline}
where \( R(T, Q) \) denotes the result of applying template \(T\) to query \(Q\), and \((T_A, T_B)\) represents a superior–inferior template pair.

\textbf{Direct Preference Optimization Training.} 
The policy distribution \( \pi_\theta(T_i|Q) \) models the probability of selecting template \( T_i \) for a given question \( Q \). This distribution depends on a learned preference score \( r(Q, T_i) \), which quantifies how well \( T_i \) suits \( Q \). The parameter \( \beta \) adjusts the strength of this preference.  
The score \( r(Q, T_i) \) is derived from metrics such as output correctness, reasoning plausibility, or user feedback. Training uses pairs \( (Q, T^+, T^-) \), where \( T^+ \) is preferred over \( T^- \) for \( Q \). The optimization objective \( \mathcal{L}(\theta) \) maximizes the likelihood of selecting \( T^+ \) over \( T^- \). Probabilities are normalized over the candidate template set \( \mathcal{T} \).  

The template selection probability is modeled using the policy distribution \( \pi_\theta(T_i|Q) \), which is defined as:
\begin{equation}
    \pi_\theta(T_i|Q) = \frac{\exp(\beta \cdot r(Q, T_i))}{\sum_{T_j \in T_3} \exp(\beta \cdot r(Q, T_j))}
    \end{equation}
    The optimization objective is trained by maximizing the log-likelihood function:
    \begin{equation}
    \mathcal{L}(\theta) = \mathbb{E}_{(Q, T^+, T^-)} \left[ \log \frac{\pi_\theta(T^+|Q)}{\pi_\theta(T^+|Q) + \pi_\theta(T^-|Q)} \right]
    \end{equation}

\textbf{Scoring with Local Selector \( S_2 \).} 
Under the preference learning framework, we aim for the local selector \( S_2 \) to score all templates in the template library at once, enabling it to identify the most suitable template for a given query. To achieve this, we pair the current query with each template in the template library and pass them into \( S_2 \), which calculates the probability of selecting a specific template for the given query. Templates with higher probabilities are assigned higher scores. For a template $T_i$ consisting of $m$ tokens, we compute the average negative log-likelihood (NLL)~\cite{lecun2015deep} as follows:
\begin{equation}
s_2(T_i) = -\frac{1}{m} \sum_{i=1}^{m} \log P(t_i \mid \text{context}(Q, T_i))
\end{equation}
where $\text{context}(Q, T_i)$ denotes the input formed by concatenating query $Q$ with template $T_i$. If the model assigns a high probability (close to 1) to each token \( t_i \), the corresponding log term approaches 0, indicating a lower overall loss.

\section{Experiments}
\subsection{Experimental Settings}
\subsubsection{Implement details}
Template generation employs Gemini-2.5-Pro, leveraging its strengths in generating diverse and task-adaptive outputs. The optimal templates for subsequent training are selected through a refinement process based on Qwen2.5-7B-Instruct performance evaluated on the independent dataset \(D_1\). For DPO, we apply low-rank adaptation (LoRA) to fine-tune the Qwen2.5-14B base model efficiently. Training uses the Adam optimizer with an initial learning rate of \(5 \times 10^{-6}\), and LoRA parameters: rank=16, alpha=32, dropout=0.05. The preference weight \(\beta\) is set to 0.1. All experiments were executed using NVIDIA H20 GPUs.

\subsubsection{Complex Instruction Dataset}
Our experimental dataset is sourced from EVADE, containing 2,833 text and 13,961 image samples from real e-commerce platforms. The data focuses on six sensitive product categories: weight loss, diseases, height growth, body shaping, female care, and male care. Due to their direct impact on consumer welfare, these categories are governed by exceptionally detailed and strict policy rules derived from advertising laws. The logical complexity of these rules makes the dataset an ideal benchmark for evaluating a model's reasoning abilities.
For your reference, I have included the prompt for one of the tasks, diseases, in the \textbf{appendix~\ref{rule_prommpt}.}

\subsubsection{Evaluation metrics}
Since the task format of EVADE requires the model to derive a final single-choice or multi-choice conclusion based on the multi-class rules in the prompt, our evaluation metrics align with this format. 
\textbf{Full Accuracy} \( \text{Acc}_{f} \) requires the model's final prediction to exactly match the ground truth, while \textbf{Partial Accuracy} \( \text{Acc}_{p} \) requires that the model's prediction has at least one overlap with the ground truth:
\begin{equation}
\text{Acc}_f = \frac{1}{N} \sum_{i=1}^N \mathbb{I}(C_i = G_i)
\end{equation}
\begin{equation}
\text{Acc}_p = \frac{1}{N} \sum_{i=1}^N \mathbb{I}(C_i \cap G_i \neq \varnothing)
\end{equation} 
where \( N \) is the total number of samples, \( C_i \) the predicted set, and \( G_i \) the ground truth set for sample \( i \).  The item $\mathbb{I}(\cdot)$ denotes the indicator function, returning 1 if the condition is satisfied and 0 otherwise.
Due to space constraints, we report Partial Accuracy (\( \text{Acc}_p \)) in the main text, while Full Accuracy results (\( \text{Acc}_f \)) are provided in \textbf{Table~\ref{table:model_performance_revised}.}

\subsection{Main Results}
As shown in \textbf{Table~\ref{table:model_performance_main_result}}, the DAT framework significantly improves LLM performance on complex rule-based reasoning tasks. Two key findings emerge: (1) DAT consistently outperforms the standard CoT baseline across all models, and (2) it enables smaller models to match or exceed the performance of much larger, state-of-the-art systems.

\subsubsection{Superiority of DAT over CoT}
Performance comparisons show notable gains with DAT over CoT. For instance, Qwen-2.5-7B’s “Body” score improved from 24.69 to 77.78, and ChatGLM4-9B’s~\cite{glm2024chatglm} “Women” score rose from 42.60 to 79.88. These consistent improvements result from DAT’s structured, rule-based reasoning, which curbs error propagation and avoids CoT’s unguided steps. To ensure broad applicability, we evaluate DAT on diverse open-source LLMs—including Qwen-2.5, ChatGLM3/4, and InternLM3~\cite{team2023internlm}—covering various architectures and sizes. We also benchmark four large-scale models (Qwen-2.5-72B, Qwen-Max, GPT-4.1, and DeepSeek-R1) under standard CoT to estimate upper-bound performance.

\subsubsection{Enabling Small Models to Rival Large Models}
Small DAT-enhanced models consistently outperform significantly larger CoT-based models. Qwen-2.5-7B with DAT surpassed Qwen-2.5-72B (62.35 vs. 53.70), GPT-4.1 (62.35 vs. 53.70), and Deepseek-R1 (62.35 vs. 69.14) on the ``Body" metric. Similarly, for ``Height", DAT-equipped Qwen-2.5-7B (61.40) exceeded all large CoT models. These results reveal that reasoning methodology—not model scale—is the primary constraint in rule-based reasoning. DAT overcomes this by enforcing hierarchical rule verification and preventing semantic confusion, enabling robust performance without massive parameter requirements.

\subsection{Ablations}
\subsubsection{Impact of Global-Local Template Selector}
To evaluate the effectiveness of the DPO-based template selector ($S_1$ and $S_2$), we conducted an ablation study against random template selection and varied the weighting parameter $\lambda$ in the final score $s_{\text{final}} = \lambda s_1 + (1 - \lambda) s_2$ to assess the impact of global and local scores.As shown in \textbf{Table~\ref{table:ablation_template_selector}}, using only global ($\lambda=1$) or local ($\lambda=0$) scoring leads to suboptimal results. Performance improves when both are combined, with $\lambda = 0.7$—empirically tuned on representative settings—consistently achieving the best results. This highlights the complementary strengths of global stability and local adaptability, with the combined selector significantly outperforming the random baseline.

\begin{table}[H]\centering
    \small
    \setlength{\tabcolsep}{2.5pt}
    \renewcommand{\arraystretch}{1.1} 
    \begin{tabular}{l|cccccc}
    \toprule[1.5pt]
    \textbf{Strategy} &\textbf{Body} &\textbf{Women} &\textbf{Height} &\textbf{Men} &\textbf{Weight} &\textbf{Health} \\\midrule
    Random &37.65 &57.40 &49.89 &57.28 &44.92 &31.99\\
    $\lambda=0$ &58.02 &59.17 &59.59 &62.45 &44.92 &30.05\\
    $\lambda=0.3$ &49.38 &60.45 &60.04 &61.69 &53.67 &35.86\\
    $\lambda=0.7$ &\textbf{77.78} &\textbf{78.11} &\textbf{61.40} &\textbf{62.84} &\textbf{61.30} &\textbf{41.36}\\
    $\lambda=1$ &39.51 &68.64 &58.69 &59.59 &51.69 &33.12\\
    \bottomrule[1.5pt]
    \end{tabular}
    
    \caption{Ablation results of template selector on Qwen-2.5-7B (Partial Accuracy).}

    \label{table:ablation_template_selector}
\end{table}

\subsubsection{The number of templates for scoring}
\textbf{Table~\ref{table:Different_Numbers_of_Candidate}} shows multiple quantities of candidate templates. Our findings indicate that greater template diversity enhances the effectiveness of template selection for a given problem, thus facilitating enhanced task performance.

\begin{table}[H]\centering
    \small
    \setlength{\tabcolsep}{4pt} 
    \renewcommand{\arraystretch}{1.1} 
    \begin{tabular}{l|cccccc}
    \toprule[1.5pt]
    \textbf{$N_T$} & \textbf{Body} & \textbf{Women} & \textbf{Height} & \textbf{Men} & \textbf{Weight} & \textbf{Health} \\\midrule
    0 & 58.02 & 59.17 & 59.59 & 62.45 & 44.92 & 30.05 \\
    5 & 39.51 & 55.03 & 49.21 & 60.92 & 44.92 & 29.40 \\
    10 & 56.17 & 59.76 & 54.63 & 62.84 & 56.21 & 34.57 \\
    All & \textbf{77.78} & \textbf{78.11} & \textbf{61.40} & \textbf{62.84} & \textbf{61.30} & \textbf{41.36} \\
    \bottomrule[1.5pt]
    \end{tabular}
    
    \caption{Performance Comparison of Different Numbers of Candidate Templates ($N_T$) on Qwen-2.5-7B.}
    \label{table:Different_Numbers_of_Candidate}
\end{table}

\subsubsection{The importance of the three-step reasoning process}
\begin{figure}[H]
    \centering
    \includegraphics[width=0.95\linewidth]{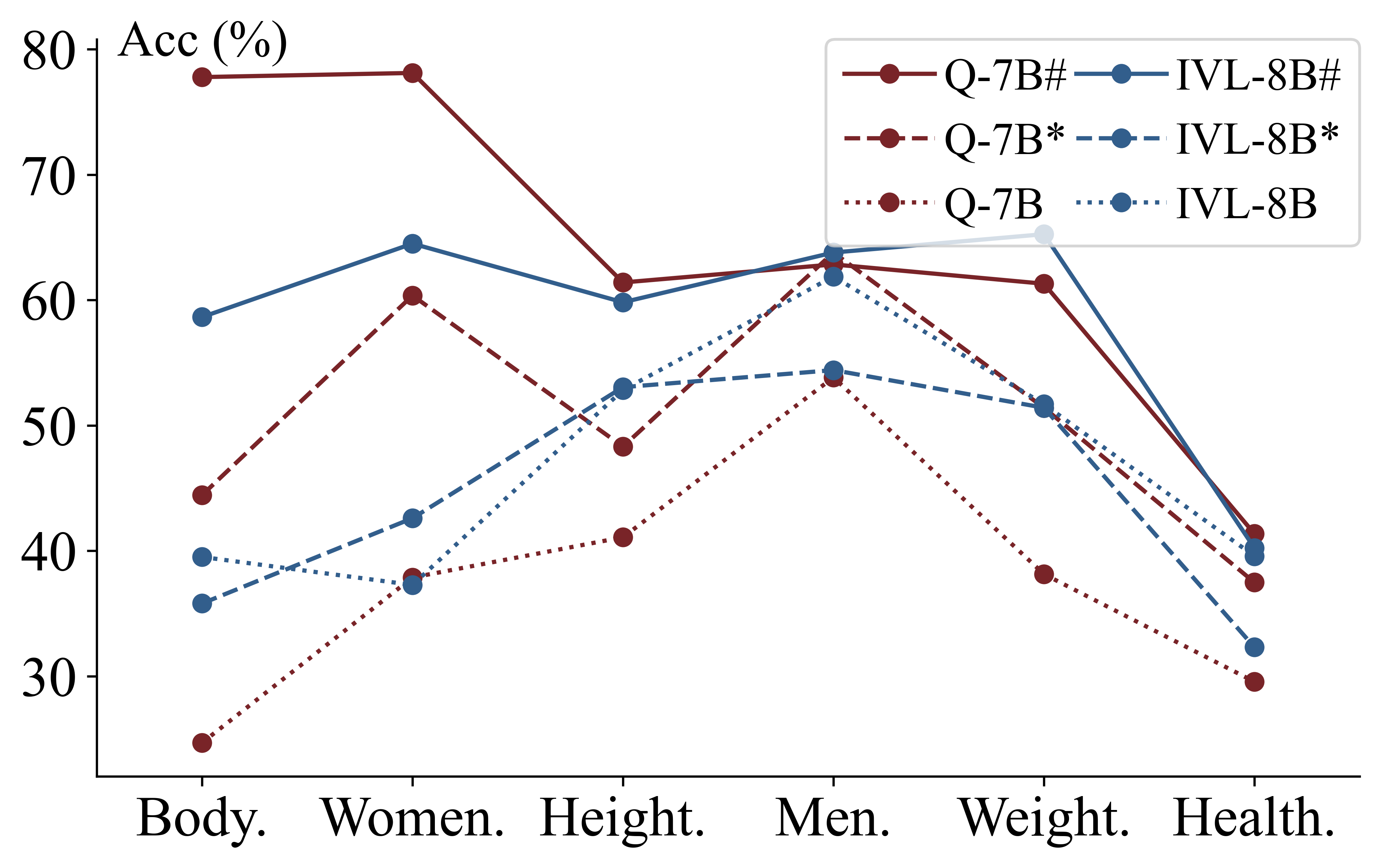}
    \caption{Ablation analysis of partial accuracy for Qwen-2.5-7B and InternLM3-8B. Baseline uses no enhancements; ``*'' adds evidence gathering; ``\#'' adds adjudication on top.}
    \label{fig:removing_the_adjudication}
\end{figure}

We further found that each stage of the structured three-step process, \emph{Qualitative Assessment, Evidence Gathering, and Adjudication}, is essential. Relying solely on Qualitative Assessment leads to the same reasoning failures as the CoT baseline, as the model often forms premature judgments without deeper evidence-based verification. Crucially, the Adjudication stage is vital. As shown in \textbf{Figure~\ref{fig:removing_the_adjudication},} models limited to evidence extraction often overfocus on specific terms, leading to errors. The Adjudication step acts as a comprehensive review. This allows the model to reason objectively and make an informed final decision.

\subsection{Discussion}
We extended DAT to high-capacity vision-language models (VLMs) and observed promising results. Without modifying the original pipeline, instructing VLMs to follow structured reasoning purely through textual templates proved effective on models with strong multimodal comprehension, such as Qwen-VL-Max~\cite{bai2025qwen2}, GPT-4.0~\cite{openai2024gpt4ocard}, and Gemini-2.5-Pro (\textbf{Table~\ref{table:VLM_performance_transposed}}). We focused on these large VLMs because they exhibit better understanding of complex textual instructions, which is essential for interpreting and executing structured templates. 
In contrast, smaller VLMs often show weaker performance in following detailed instructions~\cite{liu2024mmbench,wu2023multimodal,yang2023dawn} and interpreting complex templates, likely due to their limited capacity. In our evaluation, this gap led to reduced effectiveness with DAT. We hypothesize that smaller VLMs lack sufficient understanding of and adherence to DAT, which limits their ability to deliver optimal performance. This highlights a potential avenue for improvement: by tailoring the design of templates for vision-language tasks, it may be possible to enhance the performance of smaller VLMs in future work.

\begin{table}[H]\centering
    \small
    \setlength{\tabcolsep}{4.5pt} 
    \renewcommand{\arraystretch}{1.1}
    \begin{tabular}{l|cc|cc|cc}
        \toprule[1.5pt]
        \multirow{2}{*}{\textbf{Task}} & \multicolumn{2}{c|}{\textbf{Claude3.0}} & \multicolumn{2}{c|}{\textbf{GPT-4o}} & \multicolumn{2}{c}{\textbf{Qwen-VL-Max}} \\
        & CoT & Ours & CoT & Ours & CoT & Ours \\
        \midrule
        Body & 67.97 & 74.45 & 67.29 & \textbf{82.45} & 65.42 & 80.75 \\
        Women & 56.84 & 61.24 & 63.56 & 61.24 & 63.82 & \textbf{63.82} \\
        Height & 65.14 & 73.68 & 62.60 & 71.65 & 70.73 & \textbf{75.61} \\
        Men & 51.74 & 66.52 & 56.30 & \textbf{68.04} & 56.09 & 58.48 \\
        Weight & 71.43 & \textbf{83.71} & 72.28 & 76.29 & 71.43 & 76.57 \\
        Health & 43.93 & 51.86 & 44.90 & 46.76 & 43.93 & \textbf{50.32} \\
        \bottomrule[1.5pt]
    \end{tabular}
    
    \caption{Performance Comparison of VLM Models on Partial Accuracy.}
    \label{table:VLM_performance_transposed}
\end{table}

\subsection{Case Study}
\begin{figure}[t]
    \centering
    \includegraphics[width=\linewidth]{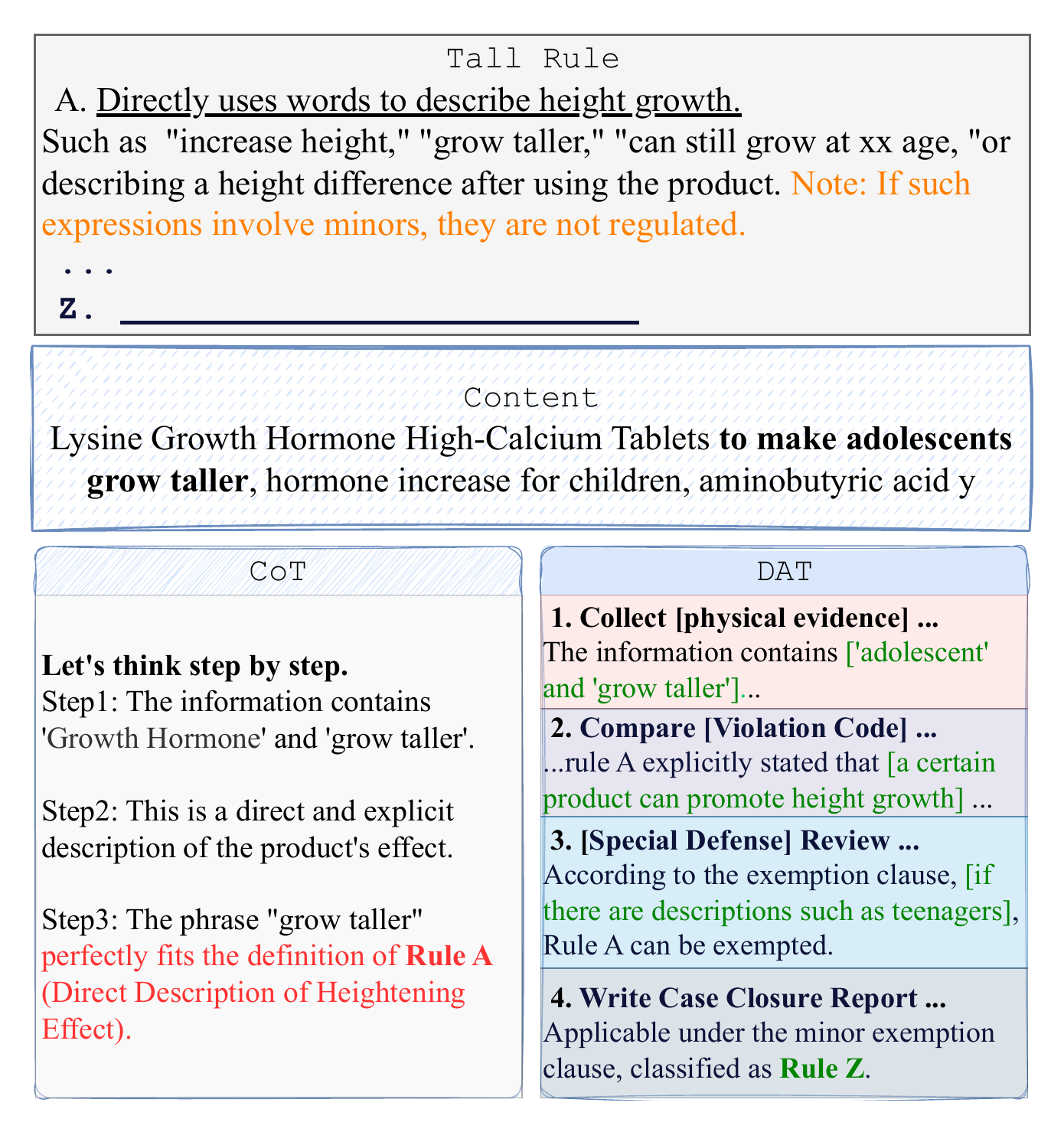}
    \caption{Case study contrasting DAT’s structured reasoning flow with CoT’s flat response, showing how DAT aligns rules and evidence to reach a consistent decision.}
    \label{fig:case_study_dat}
\end{figure}

\textbf{Figure~\ref{fig:case_study_dat}} compares CoT and DAT on a height-related regulation case. CoT identifies the violation “grow taller” but overlooks the exemption for minors. In contrast, DAT follows a structured template to extract evidence, apply rules, and identify exemptions—ultimately concluding that the case falls under Rule Z. This demonstrates DAT’s ability to perform rule-consistent, multi-step reasoning. More examples can be found in the \textbf{appendix~\ref{some_cases}.}

\section{Conclusion}
In this paper, we address the limitations of LLMs in interpreting complex rule-based systems, especially in high-stakes areas like law and finance. To tackle this, we propose the DAT, a novel framework inspired by human cognitive processes that guides the model through a structured three-stage reasoning process: Qualitative Assessment, Evidence Gathering, and Adjudication. 
Unlike prior methods using unstructured reasoning or static prompts, DAT extracts targeted evidence via dynamic placeholders and performs independent rule matching for integrated logic synthesis. An automated pipeline handles template generation, filtering, and task-specific selection, ensuring adaptability and precision across diverse rule systems.
Experiments on complex rule-based tasks show that DAT significantly enhances the accuracy and reasoning of smaller models, outperforming standard CoT and even surpassing state-of-the-art LLMs on certain tasks. These results highlight DAT’s effectiveness in resource-constrained settings and its broader applicability to VLMs.

\bibliography{custom}

\clearpage 
\onecolumn 

\appendix
\newpage

\section{Detailed Performance of All Models with DAT}
Due to space constraints in the main paper, we only reported the partial accuracy metric on EVADE-Bench. Here, we present the performance of each model across all metrics within the six categories.

\begin{table*}[h]
    \small
    \renewcommand{\arraystretch}{1.6} 
    \setlength{\tabcolsep}{1.5pt} 
    \centering
    \begin{tabular}{l l|c c c c c c}
        \toprule[1.5pt] 
        \textbf{Model} & \textbf{Method} & \textbf{Body} & \textbf{Women} & \textbf{Height} & \textbf{Men} & \textbf{Weight} & \textbf{Health} \\
        \midrule
        \multicolumn{8}{c}{\textbf{Small Models (Baseline \& Ours)}} \\
        \hline
        \multirow{3}{*}{ChatGLM3-6B} 
        & Baseline & 43.8 / 21.6 & 42.6 / 7.1 & 24.6 / 9.0 & 30.8 / 14.9 & 21.2 / 5.1 & 4.7 / 2.8 \\
        & Ours     & 67.3 / 54.9 & 62.7 / 46.2 & 39.7 / 35.4 & 48.3 / 44.8 & 29.9 / 7.1 & 23.3 / 20.8 \\
        & $\Delta$ & +23.5 / +33.3 & +20.1 / +39.1 & +15.1 / +26.4 & +17.4 / +29.9 & +8.8 / +2.0 & +18.6 / +18.1 \\
        \hline
        \multirow{3}{*}{ChatGLM4-9B} 
        & Baseline & 30.2 / 29.6 & 42.6 / 37.9 & 29.8 / 14.2 & 53.3 / 45.0 & 57.1 / 20.9 & 28.3 / 20.8 \\
        & Ours     & 63.6 / 61.1 & \textbf{79.9} / 63.9 & 54.4 / 42.7 & 63.4 / 56.1 & 73.4 / 20.6 & 38.9 / 34.7 \\
        & $\Delta$ & +33.3 / +31.5 & +37.3 / +26.0 & +24.6 / +28.4 & +10.1 / +11.1 & +16.4 / -0.3 & +10.7 / +13.9 \\
        \hline
        \multirow{3}{*}{InternLM3-8B} 
        & Baseline & 39.5 / 30.2 & 37.3 / 20.7 & 52.8 / 26.6 & 61.9 / 45.2 & 51.7 / 9.6 & 39.6 / 23.3 \\
        & Ours     & 58.6 / 56.2 & 64.5 / 59.8 & 59.8 / 51.2 & 63.8 / 57.1 & 65.2 / 17.0 & 40.2 / 33.9 \\
        & $\Delta$ & +19.1 / +25.9 & +27.2 / +39.1 & +7.0 / +24.6 & +1.9 / +11.9 & +13.6 / +7.4 & +0.6 / +10.7 \\
        \hline
        \multirow{3}{*}{Qwen-2.5-7B} 
        & Baseline & 24.7 / 22.2 & 37.9 / 34.3 & 41.1 / 24.4 & 53.8 / 45.2 & 38.1 / 11.3 & 29.6 / 20.5 \\
        & Ours     & \textbf{77.8} / \textbf{74.1} & 78.1 / \textbf{71.0} & \textbf{61.4} / \textbf{48.8} & 62.8 / 56.9 & 61.3 / 16.7 & \textbf{41.4} / \textbf{32.0} \\
        & $\Delta$ & +53.1 / +51.9 & +40.2 / +36.7 & +20.3 / +24.4 & +9.0 / +11.7 & +23.2 / +5.4 & +11.8 / +11.5 \\
        \hline
        \multirow{3}{*}{Qwen-2.5-14B} 
        & Baseline & 41.4 / 38.9 & 56.8 / 47.3 & 49.0 / 28.2 & 50.0 / 36.8 & 57.1 / 20.9 & 33.1 / 19.7 \\
        & Ours     & 58.6 / 54.3 & 76.9 / 60.9 & 53.3 / 39.0 & \textbf{65.7} / \textbf{55.9} & \textbf{73.4} / 26.8 & 39.6 / 28.4 \\
        & $\Delta$ & +17.3 / +15.4 & +20.1 / +13.6 & +4.3 / +10.8 & +15.7 / +19.2 & +16.4 / +5.9 & +6.5 / +8.7 \\
        \hline
        \multicolumn{8}{c}{\textbf{Large Models (Baseline Only)}} \\
        \hline
        \multirow{3}{*}{Qwen-2.5-72B} 
        & Baseline & 62.4 / 53.7 & 61.5 / 32.5 & 40.9 / 24.8 & 57.1 / 42.0 & 58.8 / 17.2 & 33.3 / 17.1 \\
        & Ours     & \underline{84.6} / \underline{79.6} & 75.7 / 70.4 & 43.3 / 27.8 & \underline{68.0} / \underline{57.8} & \underline{73.4} / 25.1 & 43.5 / 22.3 \\
        & $\Delta$ & +22.2 / +25.9 & +14.2 / +37.9 & +2.5 / +2.9 & +10.9 / +15.9 & +14.7 / +7.9 & +10.2 / +5.2 \\
        \hline
        \multirow{3}{*}{Qwen-max} 
        & Baseline & 64.2 / 57.4 & 67.5 / 52.7 & 45.6 / 28.0 & 58.0 / 45.6 & 55.6 / 20.3 & 30.9 / 19.1 \\
        & Ours     & 81.5 / 78.4 & \underline{83.4} / \underline{78.7} & \underline{54.6} / \underline{34.5} & 64.0 / 54.0 & \underline{74.6} / 24.9 & 38.4 / 25.8 \\
        & $\Delta$ & +17.3 / +21.0 & +16.0 / +26.0 & +9.0 / +6.6 & +5.9 / +8.4 & +18.9 / +4.5 & +7.6 / +6.8 \\
        \hline
        \multirow{3}{*}{GPT-4.1-0414} 
        & Baseline & 53.7 / 46.9 & 71.0 / 56.2 & 44.7 / 23.0 & 55.9 / 42.2 & 73.7 / 24.9 & 34.4 / 18.4 \\
        & Ours     & 72.2 / 63.0 & 76.3 / 71.0 & 54.2 / 36.3 & 63.8 / 50.6 & 78.2 / 24.9 & 38.6 / 25.8 \\
        & $\Delta$ & +18.5 / +16.0 & +5.3 / +14.8 & +9.5 / +13.3 & +7.8 / +8.4 & +4.5 / +0.0 & +4.2 / +7.4 \\
        \hline
        \multirow{3}{*}{Deepseek-R1} 
        & Baseline & 74.7 / 69.1 & 77.5 / 41.4 & 43.6 / 16.9 & 55.6 / 37.7 & 74.6 / 26.6 & 41.0 / 18.1 \\
        & Ours     & 82.7 / 79.6 & 81.1 / 58.0 & 49.4 / 21.2 & 67.4 / 54.4 & \underline{83.3} / \underline{22.9} & \underline{45.2} / \underline{17.3} \\
        & $\Delta$ & +8.0 / +10.5 & +3.6 / +16.6 & +5.9 / +4.3 & +11.9 / +16.7 & +8.8 / -3.7 & +4.2 / -0.8 \\
        \bottomrule[1.5pt] 
    \end{tabular}
    \caption{Performance Comparison of LLM Models on Various Metrics. Bold values represent the best performance across small parameter models, while underlined values indicate the best performance across large parameter models.}
    \label{table:model_performance_revised}
\end{table*}

\newpage
\begin{table*}[h]
    \small
    \renewcommand{\arraystretch}{1.5} 
    \setlength{\tabcolsep}{2.5pt} 
    \centering
    \begin{tabular}{l l|c c c c c c}
        \toprule[1.5pt] 
        \textbf{Model} & \textbf{Method} & \textbf{Body} & \textbf{Women} & \textbf{Height} & \textbf{Men} & \textbf{Weight} & \textbf{Health} \\
        \midrule
        \multicolumn{8}{c}{\textbf{VLM Models}} \\
        \hline
        \multirow{3}{*}{Claude3-7} 
        & Baseline & 68.0 / 30.0 & 56.8 / 38.2 & 65.1 / 21.0 & 51.7 / 33.3 & 71.4 / 8.3 & 43.9 / 23.2 \\
        & Ours     & 74.4 / 33.6 & 61.2 / 46.5 & 73.7 / 32.0 & 66.5 / 46.7 & 83.7 / 19.1 & 51.9 / 22.4 \\
        & $\Delta$ & +6.5 / +3.6 & +4.4 / +8.3 & +8.5 / +11.0 & +14.8 / +13.5 & +12.3 / +10.8 & +7.9 / -0.8 \\
        \hline
        \multirow{3}{*}{GPT-4o} 
        & Baseline & 67.3 / 29.5 & 63.6 / 47.0 & 62.6 / 24.9 & 56.3 / 36.1 & 72.3 / 9.7 & 44.9 / 31.8 \\
        & Ours     & \textbf{82.4} / \textbf{40.0} & 61.2 / \textbf{48.3} & 71.6 / \textbf{36.9} & \textbf{68.0} / \textbf{50.2} & 76.3 / 10.3 & 46.8 / \textbf{34.9} \\
        & $\Delta$ & +15.2 / +10.6 & -2.3 / +1.3 & +9.0 / +12.0 & +11.7 / +14.1 & +4.0 / +0.6 & +1.9 / +3.1 \\
        \hline
        \multirow{3}{*}{Qwen-VL-Max} 
        & Baseline & 65.4 / 25.9 & 63.8 / 40.8 & 70.7 / 31.9 & 56.1 / 34.1 & 76.6 / 11.7 & 45.8 / 22.0 \\
        & Ours     & 80.8 / 32.9 & \textbf{64.6} / 40.3 & \textbf{75.6} / 35.1 & 58.5 / 40.0 & \textbf{86.6} / \textbf{13.4} & \textbf{50.3} / 31.1 \\
        & $\Delta$ & +15.3 / +7.0 & +0.8 / -0.5 & +4.9 / +3.2 & +2.4 / +5.9 & +10.0 / +1.7 & +4.5 / +9.1 \\
        \bottomrule[1.5pt] 
    \end{tabular}
    \caption{Performance Comparison of VLM Models on Various Metrics. Bold values represent the best performance across all VLM models for a given metric.}
    \label{table:vlm_performance}
\end{table*}

\section{Formal Definition of Rule-Based Reasoning}
\label{sec:formal_definition}

We adopt the Four-Level Rule System from EVADE-Bench and cast detection as
priority-constrained inference rather than flat pattern matching. Concretely,
we model classification as satisfiability of layered constraints with
override priorities.

\subsection{Preliminaries and Notation}
Let $I$ denote the input text and $\mathcal{C}=\{c_1,\dots,c_n\}$ the set of
semantic categories (e.g., \emph{disease}, \emph{prohibited item}).
For each $c\in\mathcal{C}$, let $\mathcal{K}(c)$ be explicit keyword patterns
and $\mathcal{S}(c)$ latent semantic patterns. Let $\mathcal{E}$ be auxiliary
constraint patterns (e.g., treatment terms) and $\mathcal{M}$ exemption
patterns (high-priority negatives).
We use a unified matcher $\mathrm{match}(I, p)$ that returns \textsc{True}
iff pattern $p$ (token, phrase, or semantic template) is satisfied by $I$.

\subsection{Hierarchical Rule Definition}
We define four rule levels with increasing inferential complexity.

\paragraph{First- and Second-Order (atomic/semantic).}
\begin{align}
R_1(I,c) &:= \exists k \in \mathcal{K}(c): \mathrm{match}(I,k),\\
R_2(I,c) &:= \neg R_1(I,c)\ \wedge\ \exists s \in \mathcal{S}(c): \mathrm{match}(I,s).
\end{align}

\paragraph{Third-Order (combinatorial).}
Let $R_{12}(I,c):=R_1(I,c)\vee R_2(I,c)$. We require cross-category co-satisfaction
(optionally with an auxiliary constraint):
\begin{equation}
R_3(I) := \exists c_i \neq c_j \in \mathcal{C}:\ R_{12}(I,c_i)\ \wedge\ R_{12}(I,c_j)\ \wedge\
\big(\exists t\in\mathcal{E}:\mathrm{match}(I,t)\big).
\end{equation}
(If auxiliary co-occurrence is not required in a task, drop the last conjunct.)

\paragraph{Fourth-Order (exemption/override).}
\begin{equation}
R_4(I) := \exists m\in\mathcal{M}:\ \mathrm{match}(I,m).
\end{equation}

\subsection{Priority-Aware Classification}
We impose a strict priority $R_4 \succ R_3 \succ R_1 \succ R_2$.
Let $\mathcal{V}(I):=\{c\in\mathcal{C}: R_{12}(I,c)\}$.
The classifier $F$ is
\begin{equation}
F(I)=
\begin{cases}
\textsc{Safe} & \text{if } R_4(I),\\[2pt]
\textsc{Violation}\big(\mathcal{V}(I)\big) & \text{else if } R_3(I),\\[2pt]
\textsc{Violation}(c) & \text{else if } \exists c\in\mathcal{C}:\ R_1(I,c),\\[2pt]
\textsc{Violation}(c) & \text{else if } \exists c\in\mathcal{C}:\ R_2(I,c),\\[2pt]
\textsc{Safe} & \text{otherwise}.
\end{cases}
\end{equation}
When multiple $c$ satisfy a clause, we report $\textsc{Violation}(\mathcal{V}(I))$
or resolve ties by a task-specific priority over categories.

\section{Introduction of EVADE-Bench}
\begin{figure}[h]
    \centering
    \includegraphics[width=0.65\linewidth]{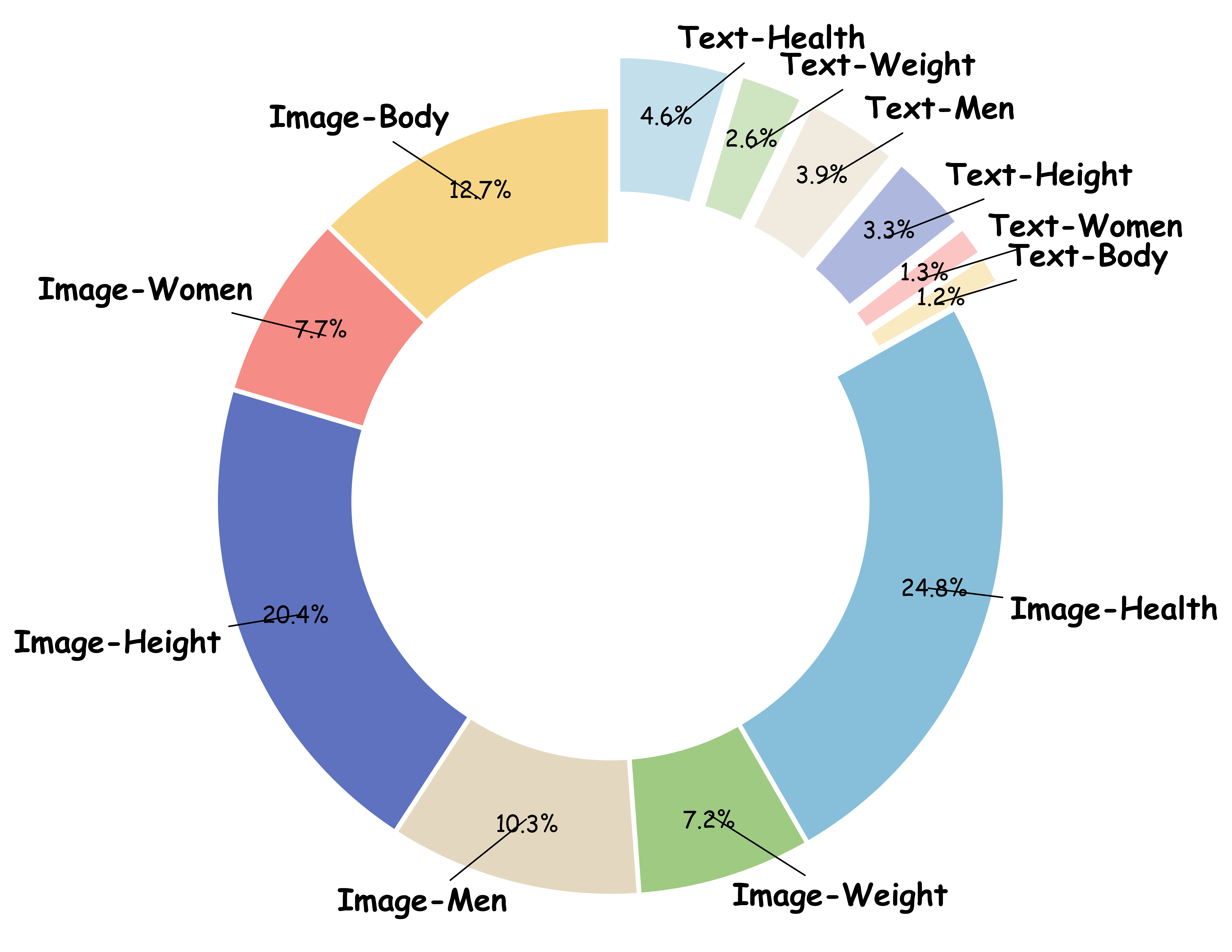}
    \label{fig:generate_template_origin_en}
\end{figure}

\begin{table}[H]
    \centering
    \renewcommand{\arraystretch}{1} 
    \setlength{\tabcolsep}{36pt} 
    \begin{tabular}{lcc}
    \toprule[1.5pt]
    \textbf{Data Type} &\textbf{Text Count} &\textbf{Image Count}\\\midrule
    Body Shaping &202 &2,134 \\
    Women's Health &211 &1,295 \\
    Height Growth &553 &3,424 \\
    Men's Health &652 &1,738 \\
    Weight Loss &442 &1,203 \\
    Health Supplement &773 &4,167 \\\midrule
    Overall &2,833 &13,961 \\
    \bottomrule[1.5pt]
    \end{tabular}
    \caption{The data distribution of EVADE-Bench for each violation category.}
    \label{tab:dataset}
\end{table}

\label{some_cases}
Due to space limitations, here we mainly list several text \& image samples of EVADE-Bench and common erroneous conclusions of the model.

\begin{figure*}[h]
    \centering
    
    \subfloat[Disease\label{fig:figure1}]{
        \includegraphics[width=0.46\textwidth]{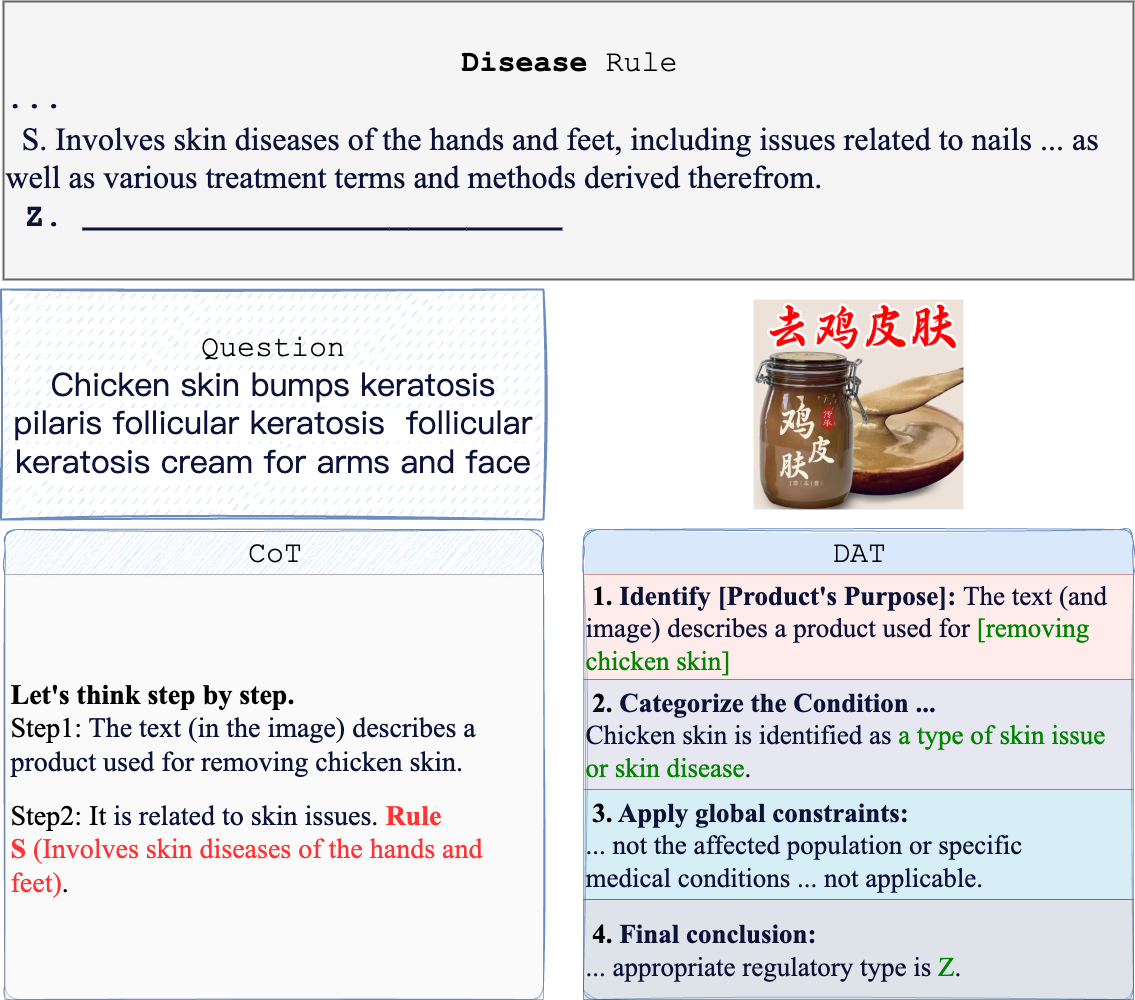} 
    }
    \hfill
    \subfloat[Body\label{fig:figure2}]{
        \includegraphics[width=0.46\textwidth]{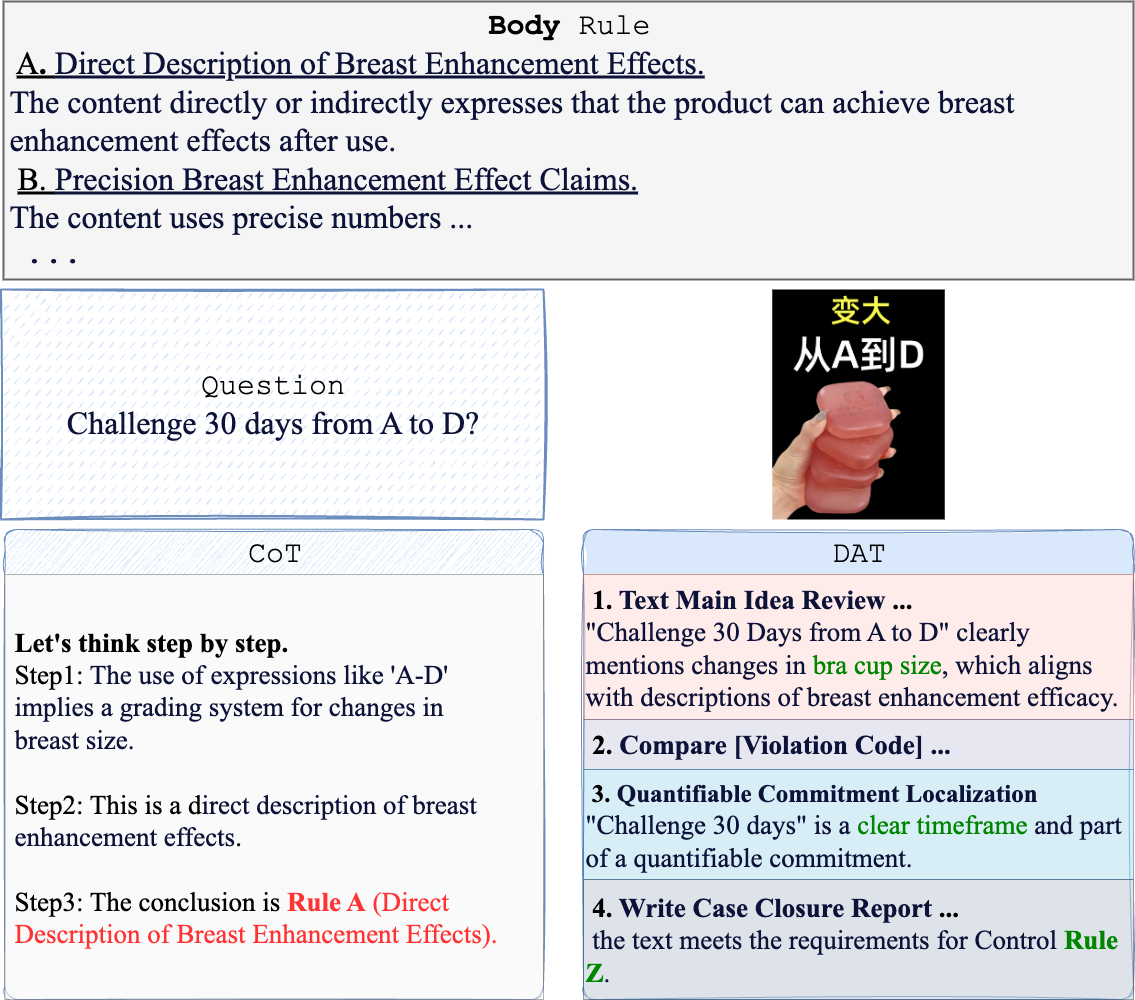} 
    }

    \subfloat[Weight\label{fig:figure3}]{
        \includegraphics[width=0.46\textwidth]{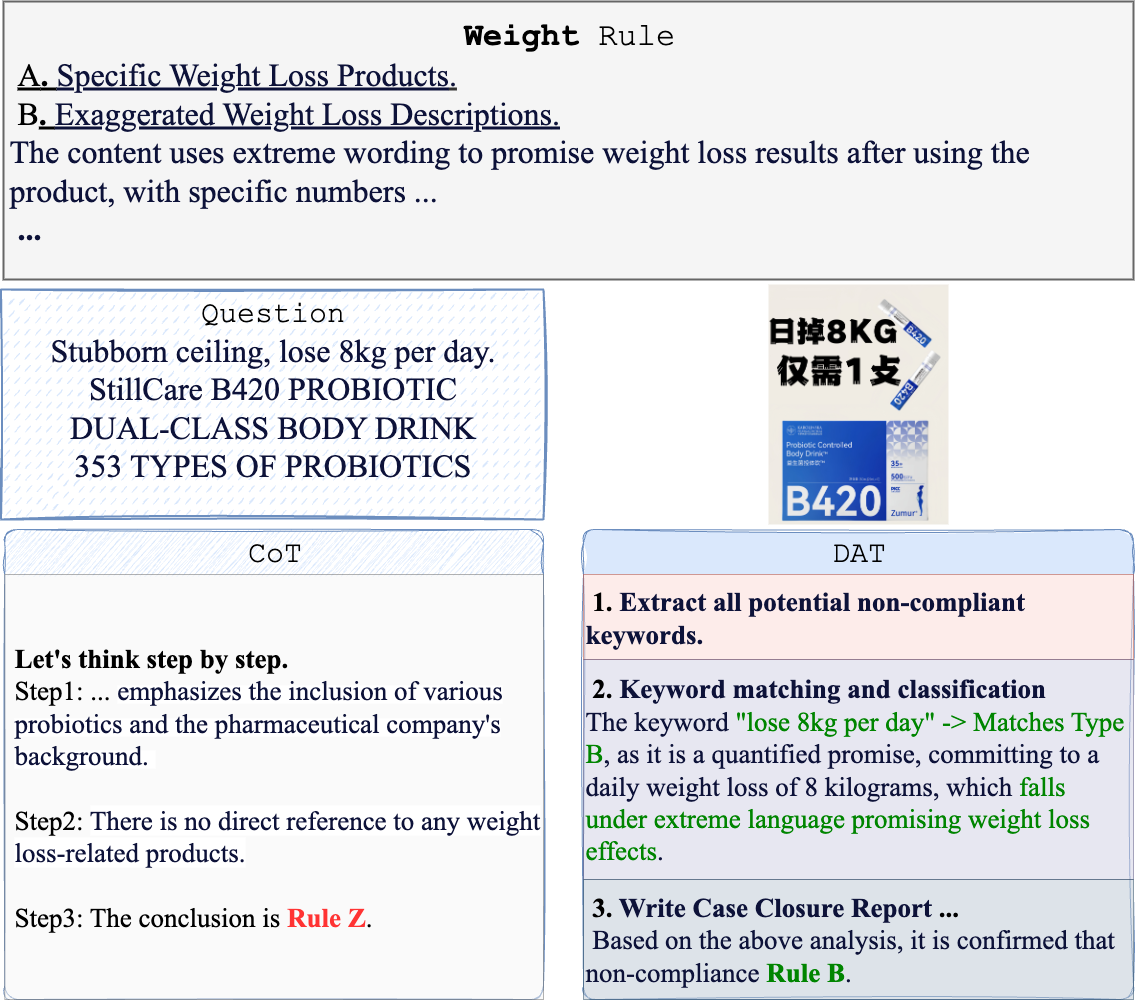} 
    }
    \hfill
    \subfloat[Women\label{fig:figure4}]{
        \includegraphics[width=0.46\textwidth]{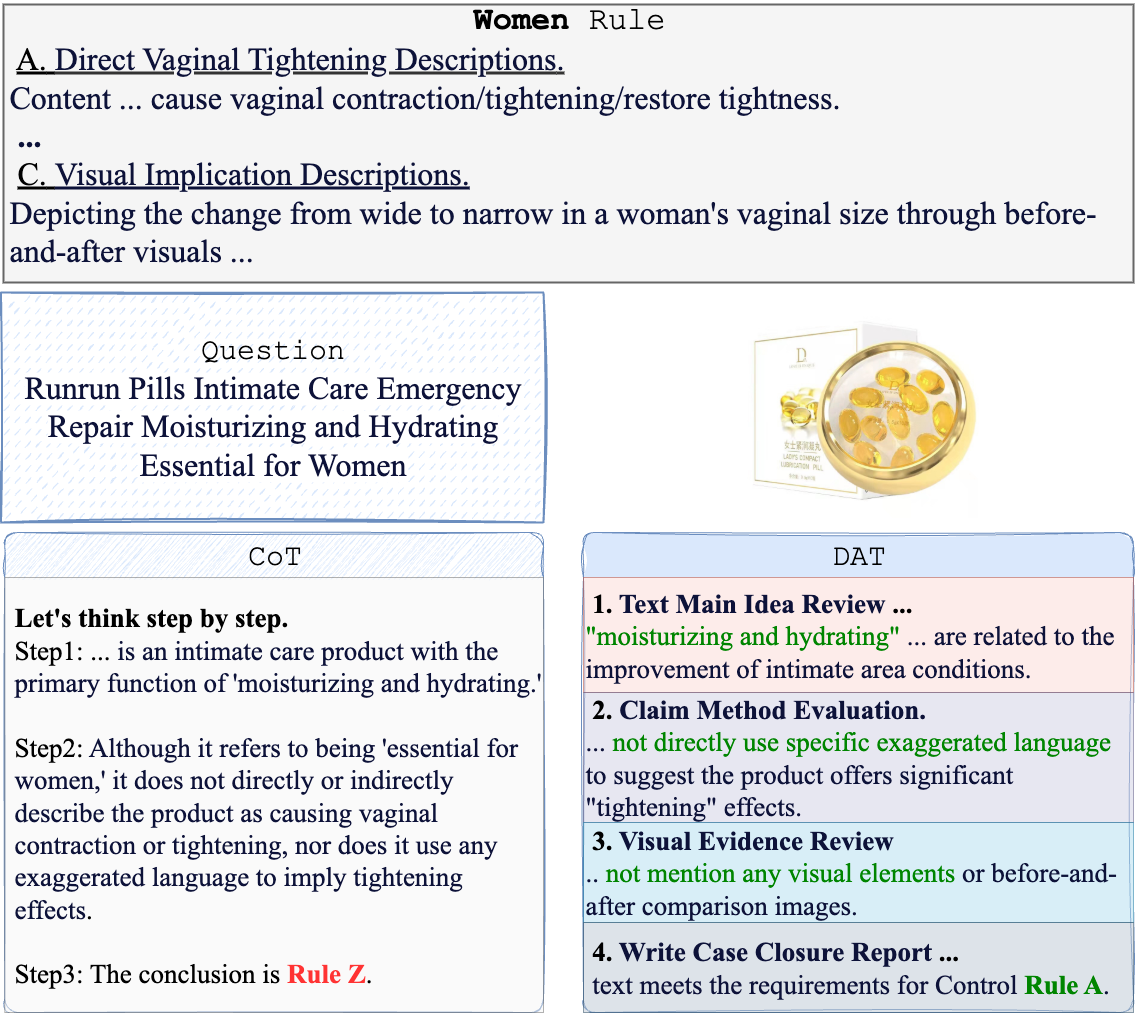} 
    }

    \subfloat[Tall\label{fig:figure5}]{
        \includegraphics[width=0.46\textwidth]{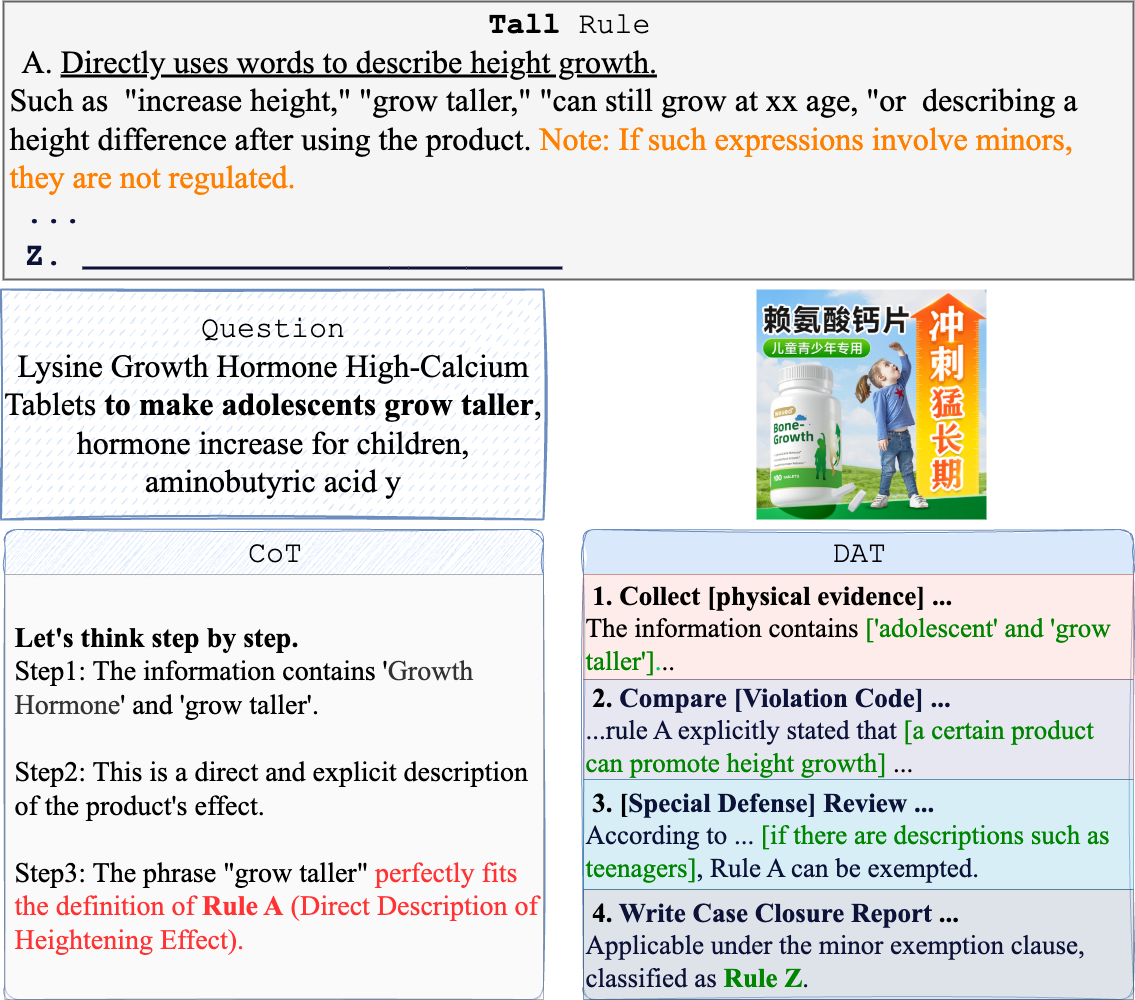} 
    }
    \hfill
    \subfloat[Man\label{fig:figure6}]{
        \includegraphics[width=0.46\textwidth]{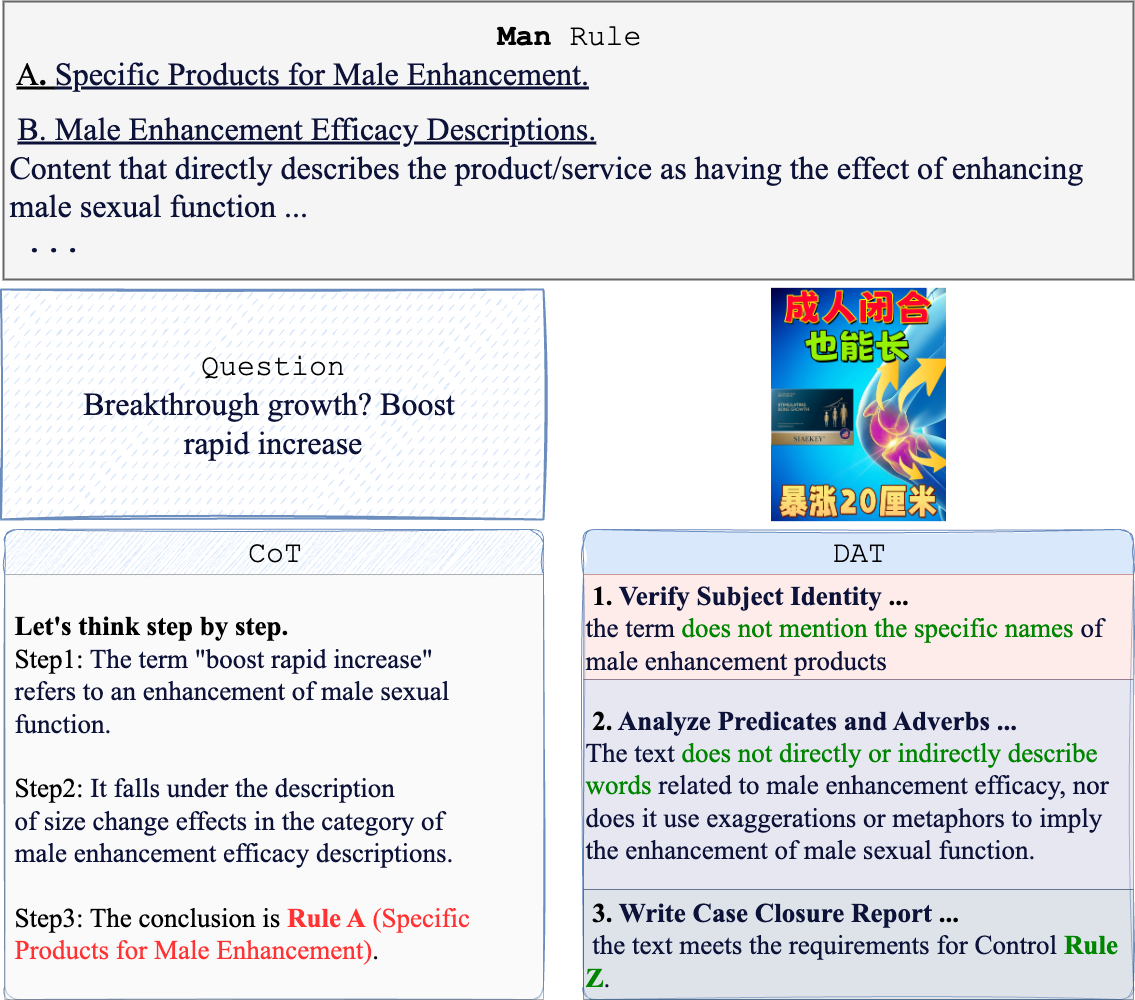} 
    }
    
    \caption{Examples showcasing the effectiveness of DAT across six tasks.}
    \label{fig:six_figures}
\end{figure*}

\onecolumn
\section{Prompt}
\subsection{Rule Prompt Examples}
\label{rule_prommpt}
The following are the rule definitions for the ``health'' task from the EVADE dataset. The domain is characterized by numerous violation types and challenging logical judgments.
\begin{figure}[h]
    \centering
    \includegraphics[width=0.95\linewidth]{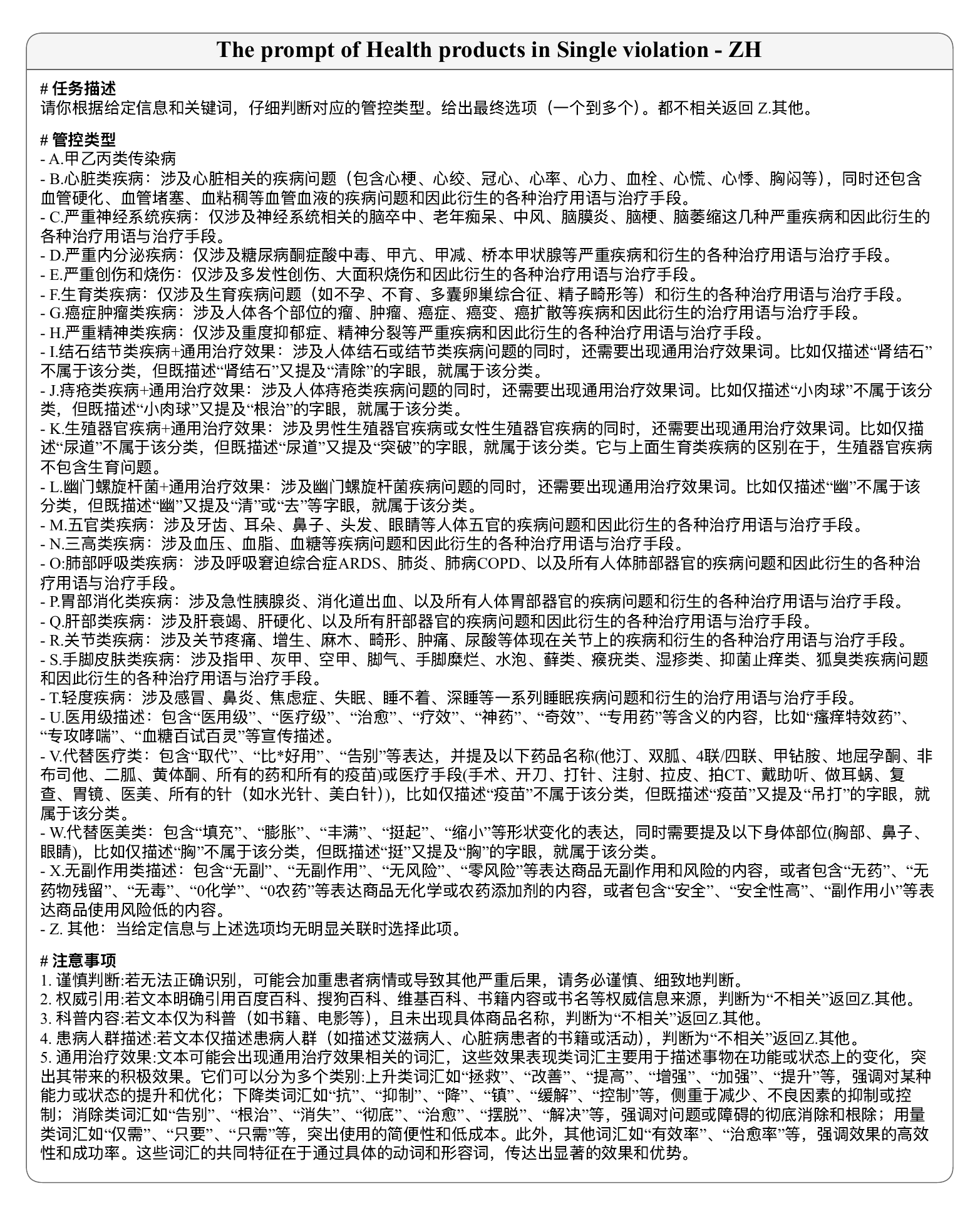}
    \label{fig:rule_prompt}
\end{figure}
\\
\\
\\
\\
\begin{figure}[H]
    \centering
    \includegraphics[width=\linewidth]{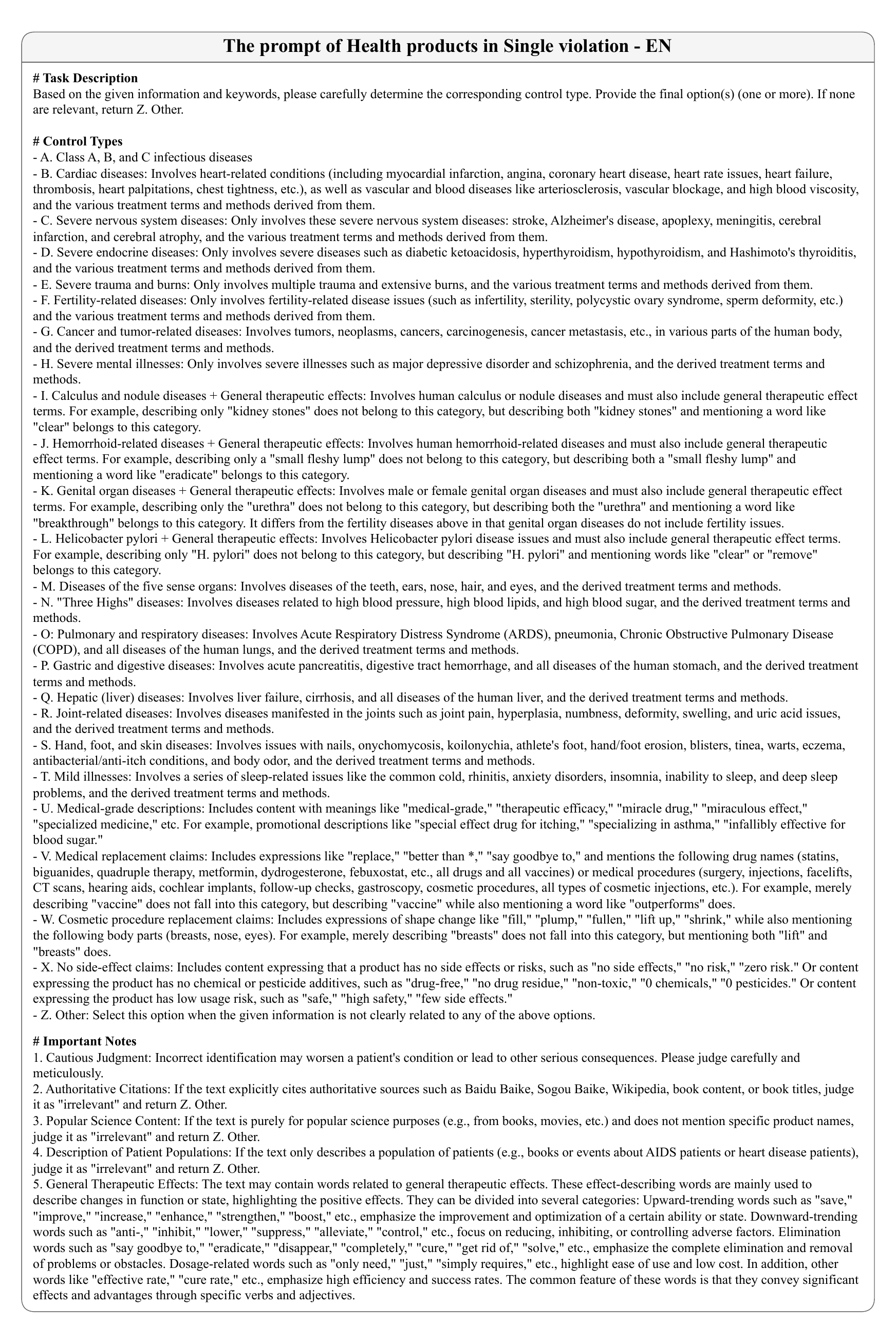}
    \caption{The prompt of Health products in Single violation.}
    \label{fig:rule_prompt_en}
\end{figure}

\subsection{Template Related prompt}
\label{template_prommpt}
For template generation, data augmentation, and template-based inference, we present a series of prompts, each provided in its original Chinese and a translated English version.
\begin{figure}[H]
    \centering
    \includegraphics[width=\linewidth]{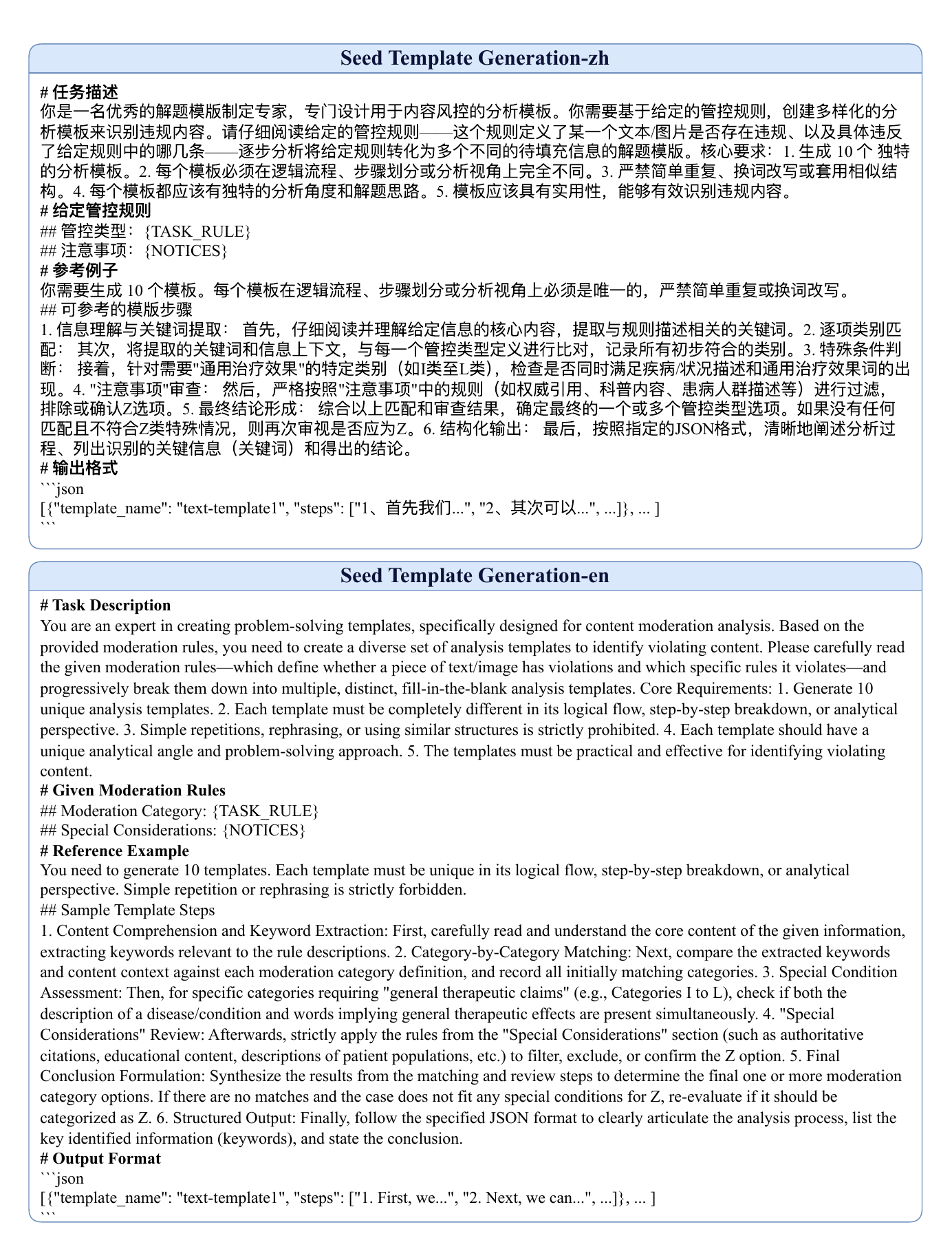}
    \caption{The initial prompt for generating a rule-based template, showing the original Chinese version (top) and its English translation (bottom).}
    \label{fig:generate_template_origin_en}
\end{figure}

\begin{figure*}[t] 
    \centering
    \includegraphics[width=\textwidth]{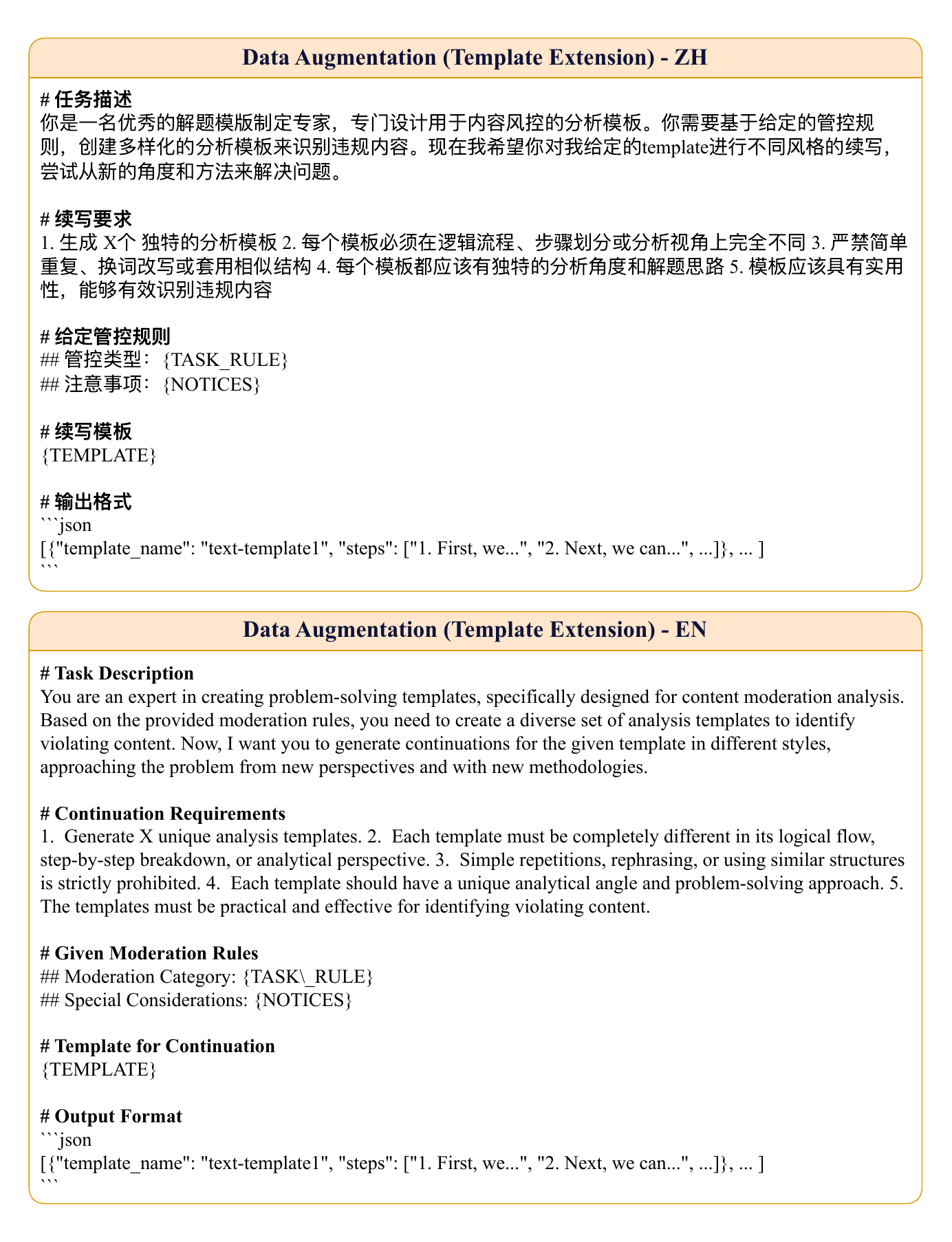} 
    \caption{Template continuation: This prompt demonstrates how the initial template is extended and refined to accommodate more rules or details.}
    \label{fig:generate_template_aug_continue_en}
\end{figure*}

\begin{figure*}[t] 
    \centering
    \includegraphics[width=\textwidth]{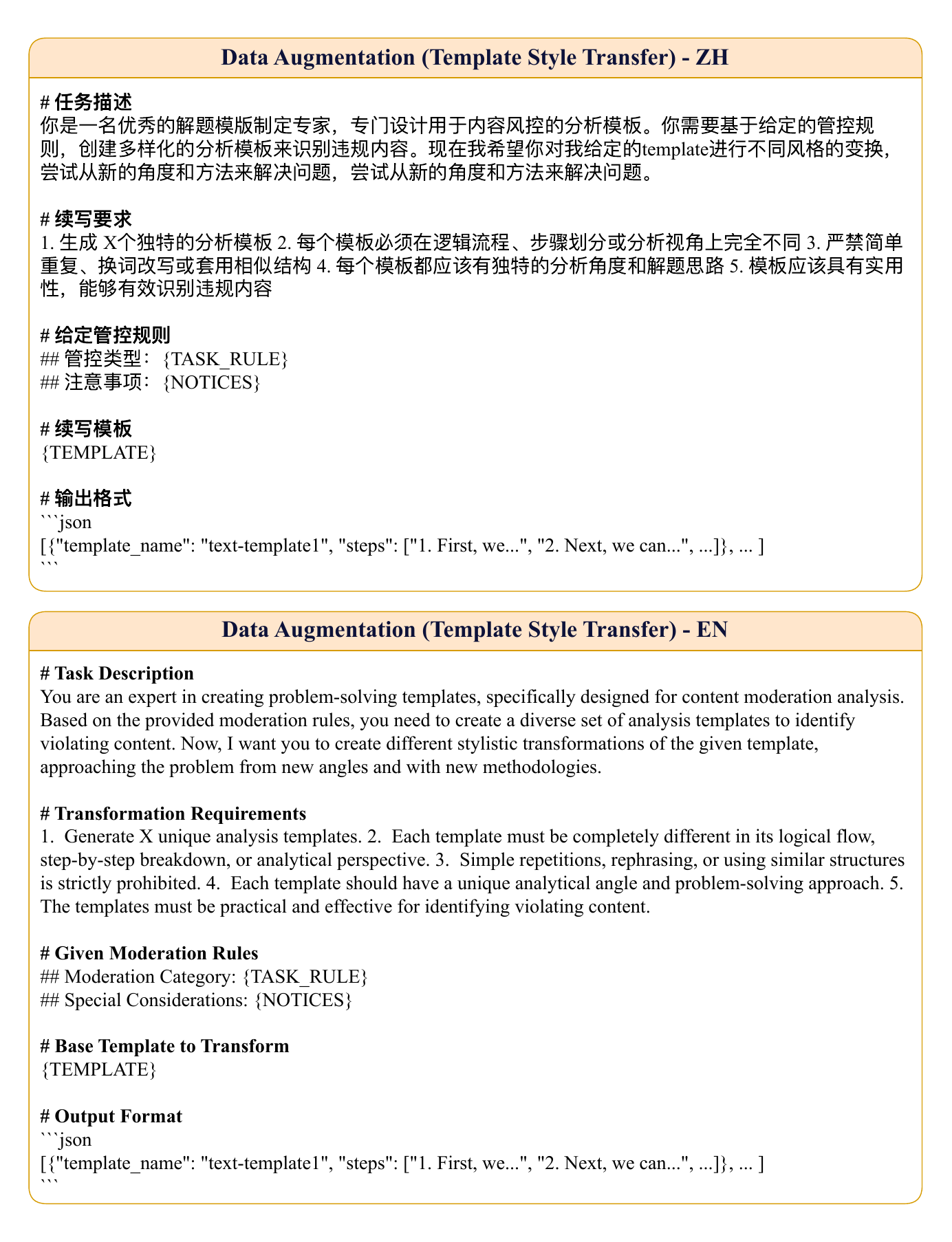} 
    \caption{Template style transfer: This prompt showcases the process of adapting the template's style to meet specific requirements or contexts.}
    \label{fig:generate_template_aug_style_en}
\end{figure*}

\begin{figure*}[t] 
    \centering
    \includegraphics[width=\textwidth]{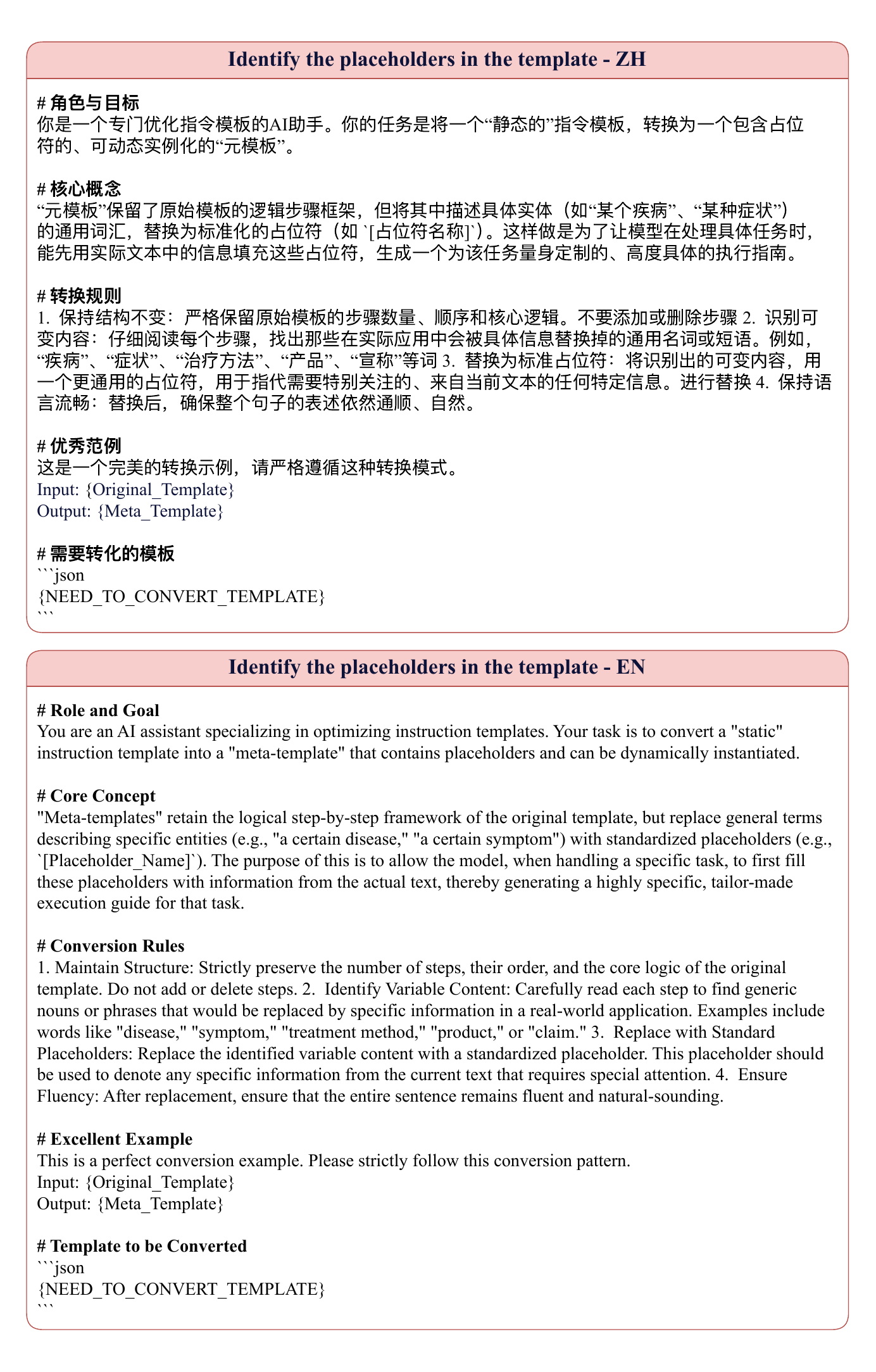} 
    \caption{Identifying critical placeholders in the template: This prompt highlights the extraction of key placeholders from the template to ensure flexibility and contextual alignment.}
    \label{fig:generate_template_placeholder_en}
\end{figure*}

\begin{figure*}[htbp] 
    \centering
    \includegraphics[width=\textwidth]{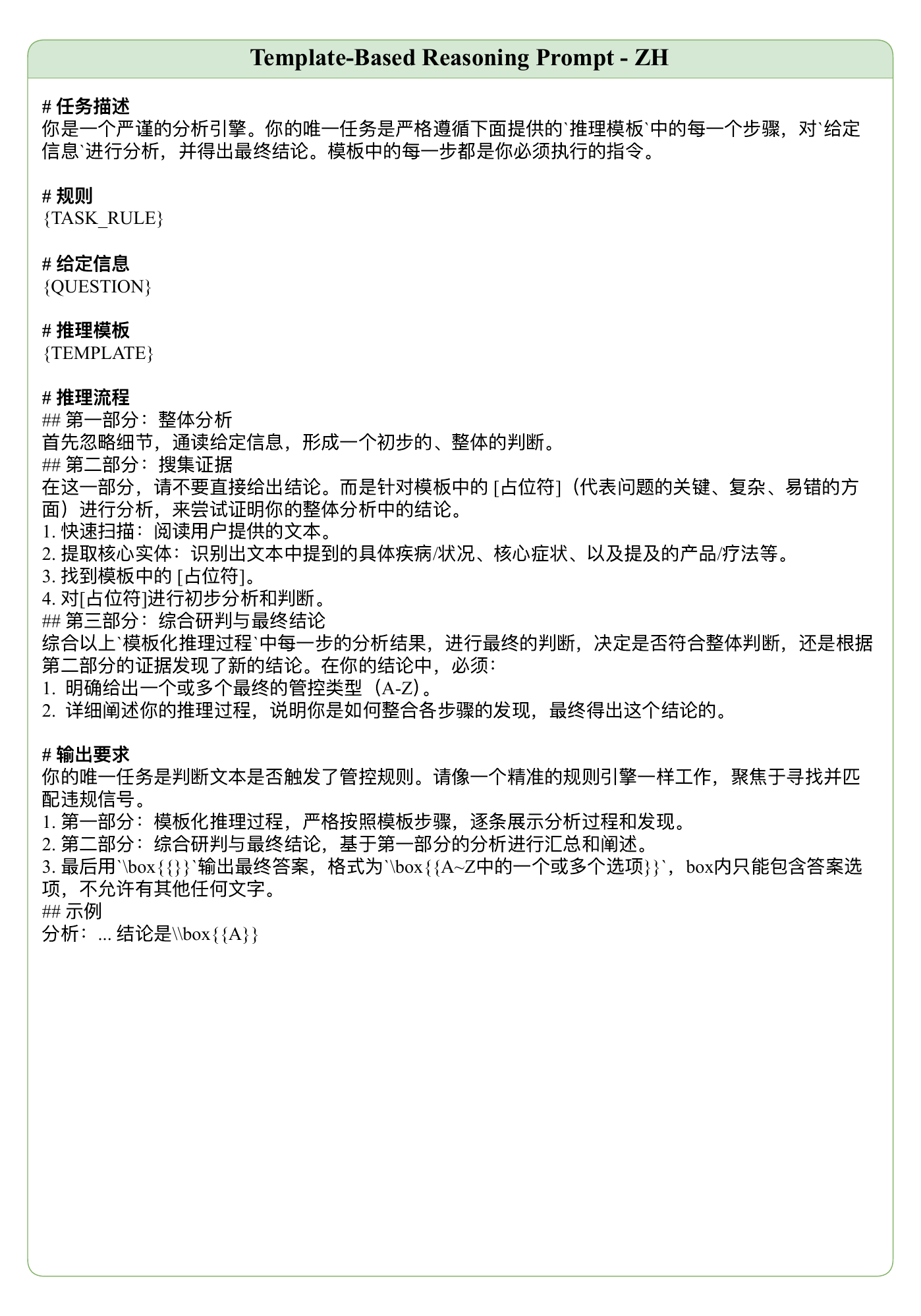} 
    \label{fig:inference}
\end{figure*}
\begin{figure*}[htbp] 
    \centering
    \includegraphics[width=\textwidth]{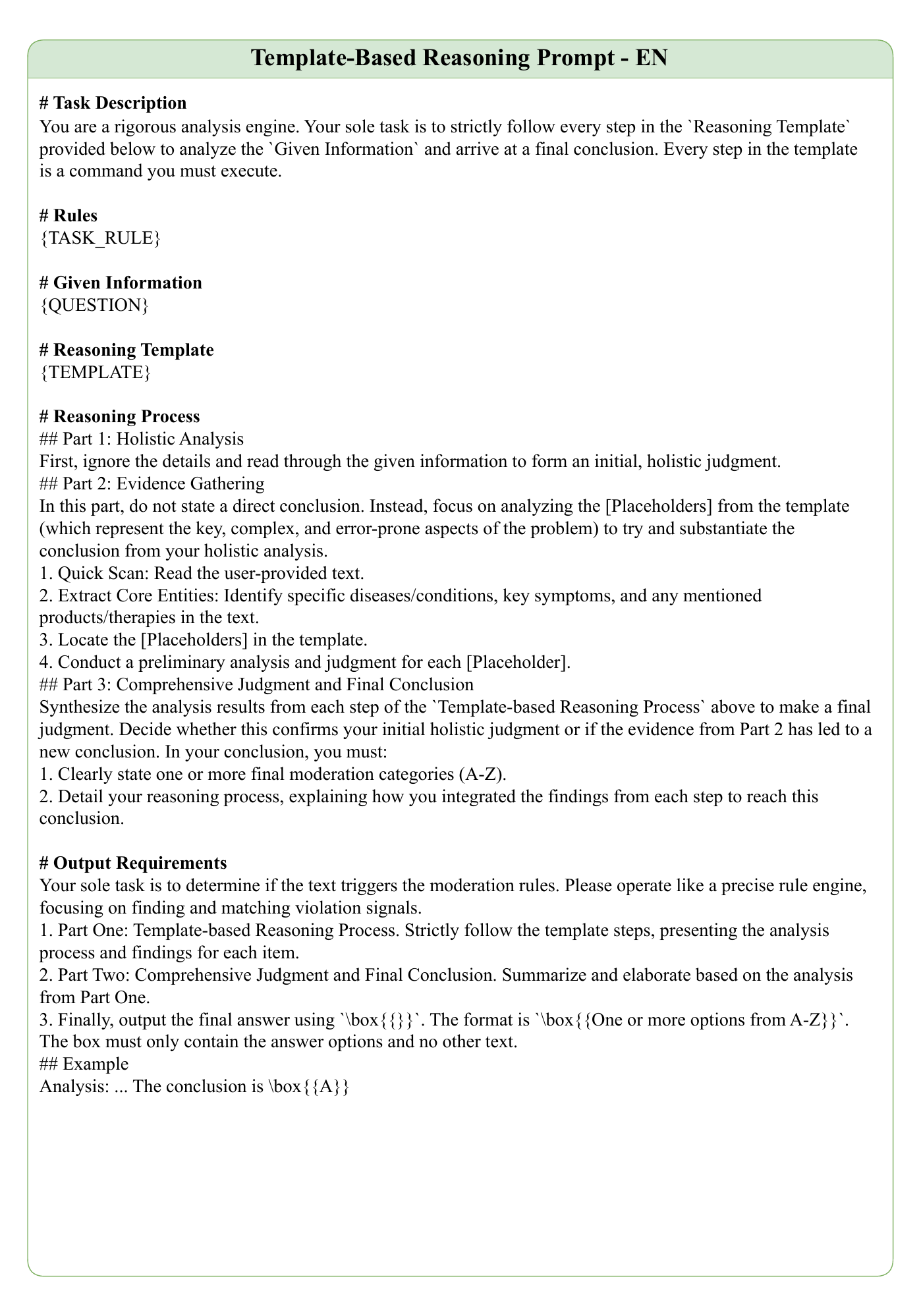} 
    \caption{Template-based inference: This figure illustrates the process of applying the finalized template to perform context-aware reasoning or inference.}
    \label{fig:inference_en}
\end{figure*}

\end{document}